\documentclass{article}

\usepackage{microtype}
\usepackage{graphicx}
\usepackage{subfigure}
\usepackage{booktabs} 

\usepackage{hyperref}

\usepackage[accepted]{icml2023}

\usepackage{amsmath}
\usepackage{amssymb}
\usepackage{mathtools}
\usepackage{amsthm}

\usepackage[capitalize,noabbrev]{cleveref}

\theoremstyle{plain}

\theoremstyle{definition}

\theoremstyle{remark}

\usepackage[textsize=tiny]{todonotes}

\def \name{P2C}
\newcommand{\ms}[1]{\tiny{$\pm$#1}}
\definecolor{revision}{rgb}{0, 0.7, 0.3} 

\definecolor{Gray}{gray}{0.9}
\definecolor{Blue9}{rgb}{0.098,0.3,0.9}
\definecolor{Red7}{rgb}{0.941, 0.243, 0.243}
\definecolor{Green7}{RGB}{55, 178, 77}
\definecolor{BrickRed}{rgb}{0.6,0,0}
\definecolor{RoyalBlue}{rgb}{0,0,0.8}
\definecolor{Tdgreen}{rgb}{0,0.4,0.7}
\definecolor{CYRed}{RGB}{228, 0, 43}
\definecolor{CYPurple}{RGB}{215, 153, 93}
\definecolor{DarkBlue}{rgb}{0,0.08,0.45}

\hypersetup{
colorlinks=true,
linkcolor=RoyalBlue,
citecolor=RoyalBlue,
filecolor=RoyalBlue,
urlcolor=RoyalBlue}

\usepackage{adjustbox}
\usepackage{multirow}
\usepackage{multicol}

\usepackage{graphicx}
\usepackage{subfigure}
\usepackage{amssymb}
\usepackage{tabularx}

\usepackage{url}
\usepackage{subfigure}
\usepackage{wrapfig}
\usepackage{amsmath}
\usepackage{amsthm}
\usepackage{bbm}

\icmltitlerunning{Prefer to Classify: Improving Text Classifiers via Auxiliary Preference Learning}

\begin{document}

\twocolumn[
\icmltitle{Prefer to Classify: Improving Text Classifiers via Auxiliary Preference Learning}




\begin{icmlauthorlist}
\icmlauthor{Jaehyung Kim}{kaist,visit}
\icmlauthor{Jinwoo Shin}{kaist}
\icmlauthor{Dongyeop Kang}{umn}
\end{icmlauthorlist}

\icmlaffiliation{kaist}{KAIST}
\icmlaffiliation{visit}{Work was mainly done while visiting Minnesota NLP.}
\icmlaffiliation{umn}{University of Minnesota}

\icmlcorrespondingauthor{Jaehyung Kim}{jaehyungkim@kaist.ac.kr}

\icmlkeywords{Machine Learning, ICML, NLP, Preference Learning}

\vskip 0.3in
]

\printAffiliationsAndNotice{}  

\begin{abstract}

The development of largely human-annotated benchmarks has driven the success of deep neural networks in various NLP tasks. 
To enhance the effectiveness of existing benchmarks, collecting new additional input-output pairs is often too costly and challenging, particularly considering their marginal impact on improving the current model accuracy. 
Instead, additional or complementary annotations on the existing input texts in the benchmarks can be preferable as an efficient way to pay the additional human cost. 
In this paper, we investigate task-specific preferences between pairs of input texts as a new alternative way for such auxiliary data annotation.
From `pair-wise' comparisons with respect to the task, the auxiliary preference learning enables the model to learn an additional informative training signal that cannot be captured with `instance-wise' task labels.
To this end, we propose a novel multi-task learning framework, called prefer-to-classify (P2C), which can enjoy the cooperative effect of learning both the given classification task and the auxiliary preferences.
Here, we provide three different ways to collect preference signals in practice: (a) \textit{implicitly} extracting from annotation records (for free, but often unavailable), (b) collecting \textit{explicitly} from crowd workers (high paid), or (c) pre-trained large language models such as GPT-3 (low paid). 
Given existing classification NLP benchmarks, we demonstrate that the proposed auxiliary preference learning via \name{} on them is effective in improving text classifiers. {Our codes are publicly available.\footnote{\url{https://github.com/minnesotanlp/p2c}}}

\end{abstract}

\section{Introduction}

The recent development of natural language processing (NLP) systems significantly boosts state-of-the-art performances on various NLP tasks \cite{brown2020language, ouyang2022training}.
This success of NLP systems has been driven by, among other things, the construction of largely human-annotated benchmarks, such as GLUE \cite{wang2019glue}, SQuAD \cite{rajpurkar2016squad}, or BIG-bench \cite{srivastava2022beyond}. 
These benchmarks are usually constructed by (a) collecting (or writing) the relevant input texts and (b) assigning output labels by human annotators.
Here, (a) is arguably more costly and cumbersome in many practical scenarios; for example, input texts with distribution mismatch or spurious patterns could make the model suffer from learning the generalized representation \cite{gururangan2018annotation, karamcheti2021mind}, and hence the much higher cost is often paid to the collection process to keep the quality of the constructed benchmark \cite{kaushik2020learning}. 
Therefore, it is often preferable to pay the additional human cost to annotate the existing benchmarks in a complementary way (instead of collecting new input texts), \textit{e.g.}, one can improve the label quality \cite{nie2020can, fornaciari2021beyond} by assigning multiple annotators to each input or obtain the finer task information with the new label space \cite{williams2022anlizing}.
In this paper, we investigate a new alternative way to \textit{better exploit the existing benchmarks (input texts and task labels), with auxiliary annotation} to further improve the model performance.

\begin{figure*}[t]
\begin{center}
    {
    \subfigure[Pair-wise preference signals]
        {
        \includegraphics[width=0.3\textwidth]{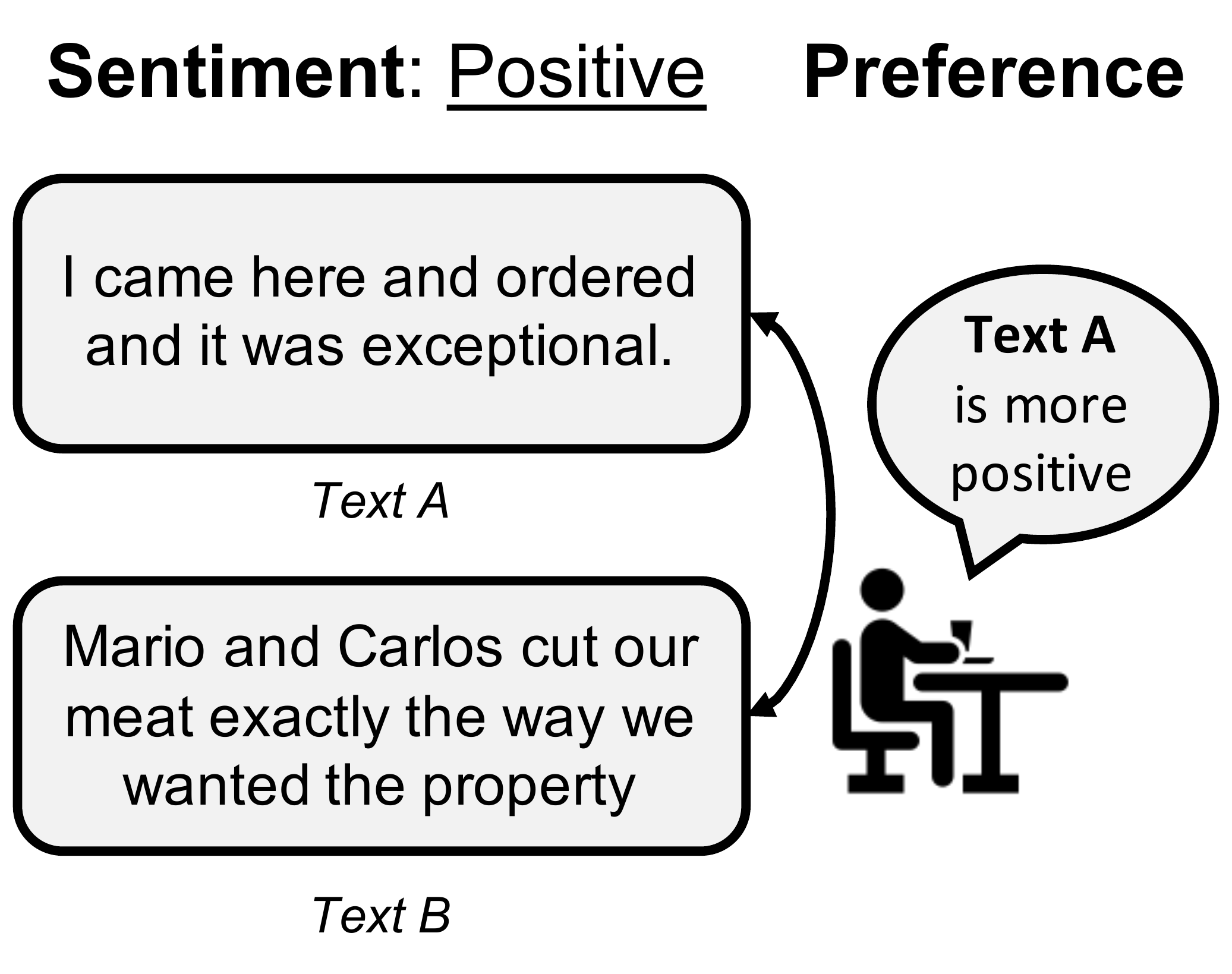}
        \label{fig:1a}
        }
    \subfigure[Alignment to human annotations]
        {
        \includegraphics[width=0.3\textwidth]{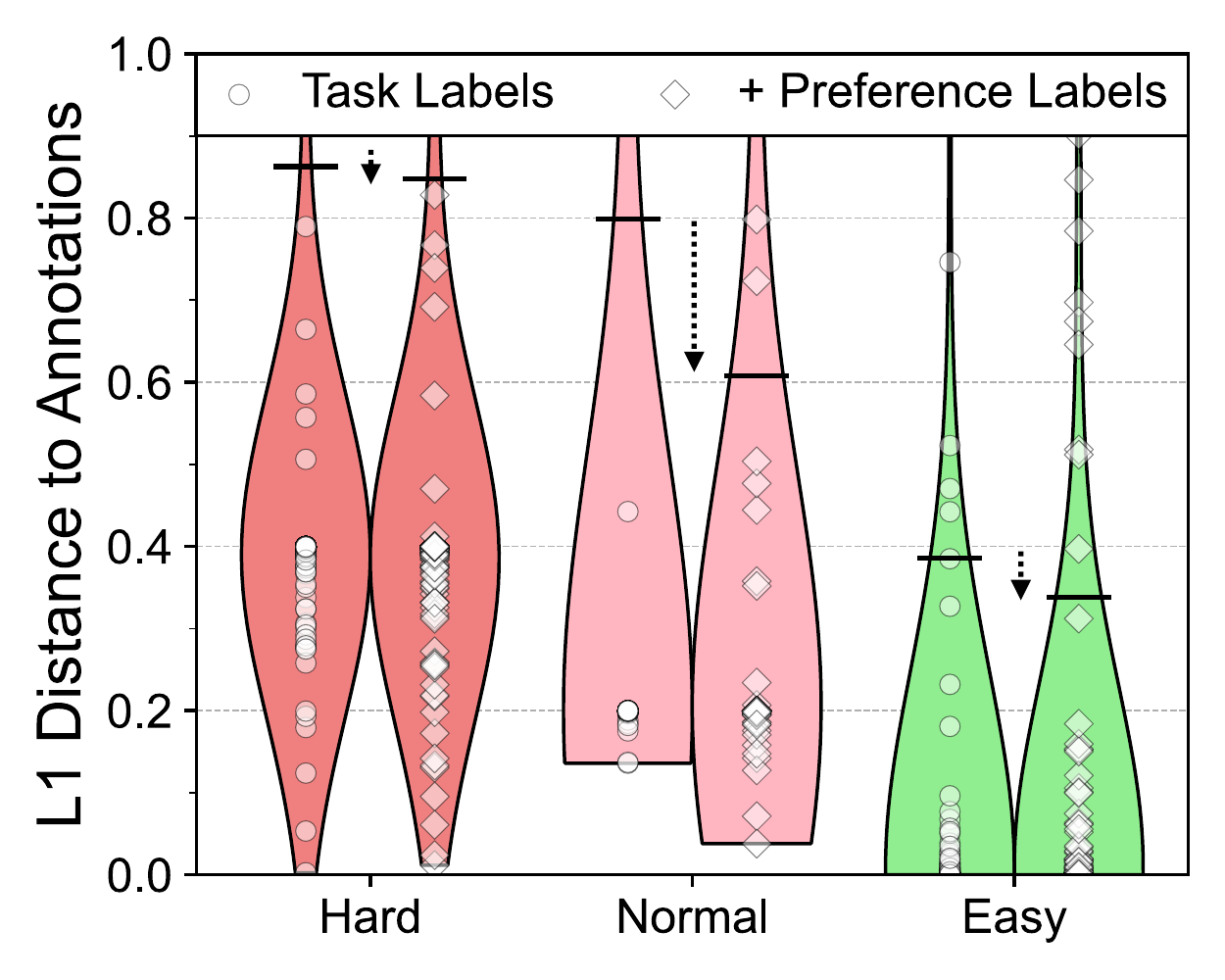}
        \label{fig:1b}
        }
    \subfigure[Improved text classification]
        {
        \includegraphics[width=0.3\textwidth]{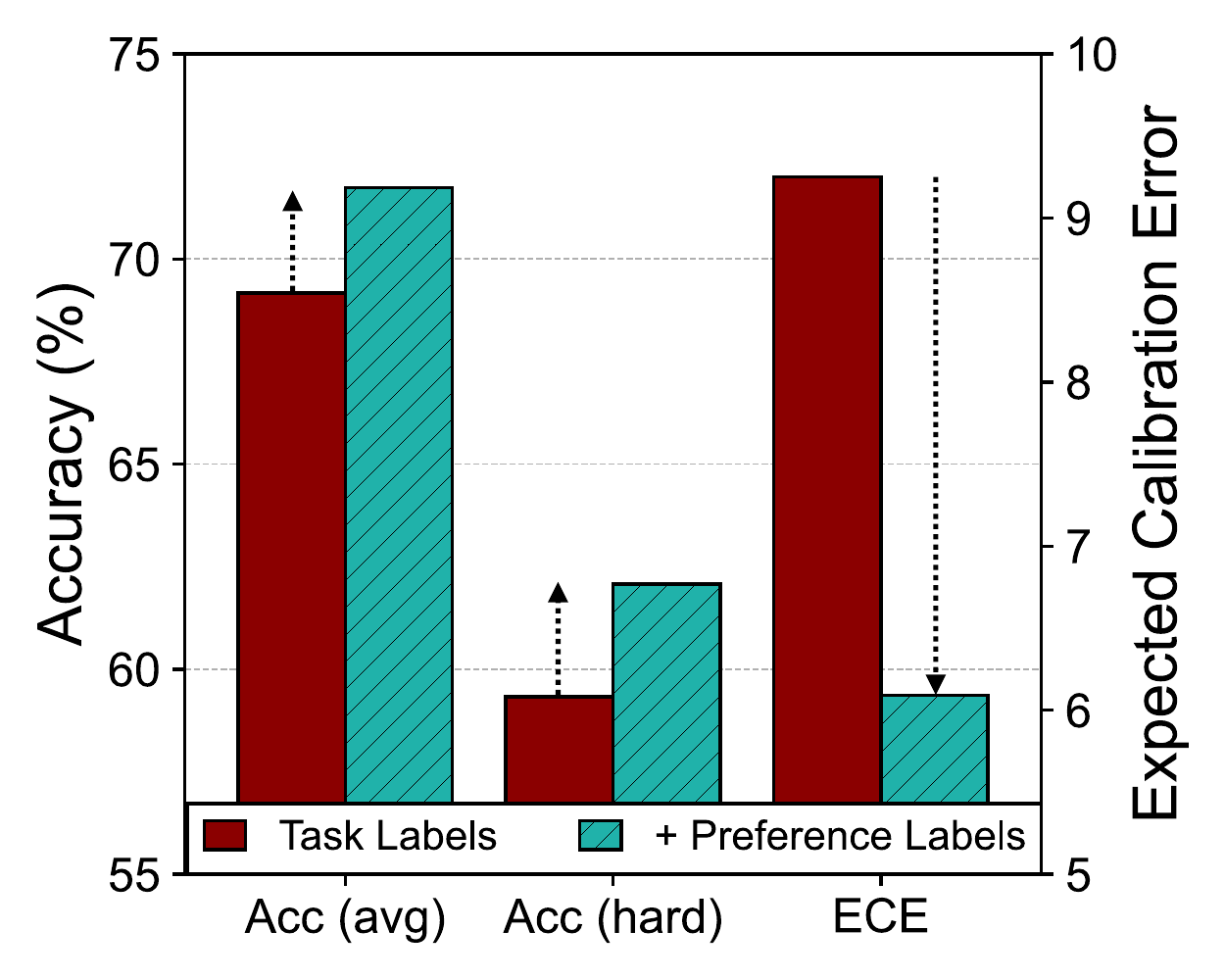}
        \label{fig:1c}
        } 
    }
\end{center}
\vspace{-0.1in}
\caption{(a) Example of a pair-wise preference signal in the sentiment classification. (b) Auxiliary preference learning makes the classifier capture the fine-grained task information; \textit{e.g.}, predictions of the classifier become more aligned with human annotations. Test samples are divided into Hard, Normal, and Easy based on the annotators' disagreement. (c) Improvement from the collected preference and \name{} in various aspects, \textit{e.g.,} better accuracy and calibration. More results are presented in Section \ref{section:experiments_subject}. 
\vspace{-0.1in}
}
\label{fig:concept}
\end{figure*}
\textbf{Contribution.} 
We introduce task-specific preferences between pairs of input texts as a new and auxiliary data annotation, to improve the text classification system upon the existing task annotations (Figure \ref{fig:1a}).
By relatively ordering a pair of two texts and better calibrating them with respect to the task through `pair-wise' comparison, we expect that the auxiliary preference learning provides an additional informative training signal that cannot be captured with `instance-wise' evaluation (see Figure \ref{fig:1b}). 

This preference signal could be obtained not only from human annotators (called \textit{subjective} preference),
but also from the existing annotation records (called \textit{extractive} preference), if available, and even the recent strong pre-trained language models (called \textit{generative} preference). 
To be specific, generative preference is obtained by querying the preference between two sentences to the recent language models (LMs), \textit{e.g.}, GPT-3 \cite{brown2020language}, with a prompting.
Next, extractive preference is constructed from the existing annotation records in datasets `without additional cost'; if one sample has been less voted than the other, we treat the latter as a relatively higher preference between the two samples.
Finally, we collect subjective preferences for 5,000 pairs of texts from (paid) crowd workers by asking them which text is more preferred to the task label.

To utilize both existing class labels and newly obtained preference labels with their cooperative effect, 
we propose a novel multi-task learning framework, coined \textit{prefer-to-classify} (\name{}), 
to effectively train the model from both classification and preference learning tasks. 
Specifically, we first introduce diverse multiple preference heads beside the classification head of the model for better learning from preference labels. 
Then, we introduce a new consistency regularization between classification and preference heads for imposing the model to have higher classification confidence in the preferred samples and hence enabling to detection of the inherent relationship between two tasks.
Lastly, we propose two advanced sampling schemes to select more informative text pairs for improving the efficiency of training.

Through the extensive experiments on ten text classification datasets, we demonstrate the effectiveness of our new auxiliary preference learning framework via \name{}; for example, \name{} with generative preference from GPT-3 exhibited 11.55\% relative test error reduction on average compared to the standard training method of the classifier.
Next, \name{} with extractive preference even outperforms the state-of-the-art methods utilizing annotation records with the 4.27\% relative test error reduction.
Lastly, the newly-collected subjective preference labels show the largest improvement compared to generative and extractive ones, which reveals the benefit of a more accurate preference signal; it does not only with the improvement in task performance but also with better calibration and task modeling; for example, 6.09\% of expected calibration error while 9.19\% from the same number of task labels.
Overall, our work highlights the effectiveness of preference learning as an auxiliary method to improve the classification system, and we believe our work could inspire researchers to consider a new alternative way for data annotation. 
\section{Improving Text Classifiers via Auxiliary Preference Learning}\label{sec:method}

In this section, we present prefer-to-classify (\name{}), a novel multi-task learning framework to use the preference labels as an auxiliary data annotation for improving the text classifier.
The auxiliary preference learning via \name{} could provide a new informative training signal that cannot be captured with the existing ‘instance-wise’ evaluation by relatively ordering a pair of two texts and better calibrating them with respect to the task through ‘pair-wise’ comparison.

\begin{figure*}[t]
\begin{center}
    \includegraphics[width=0.9\textwidth]{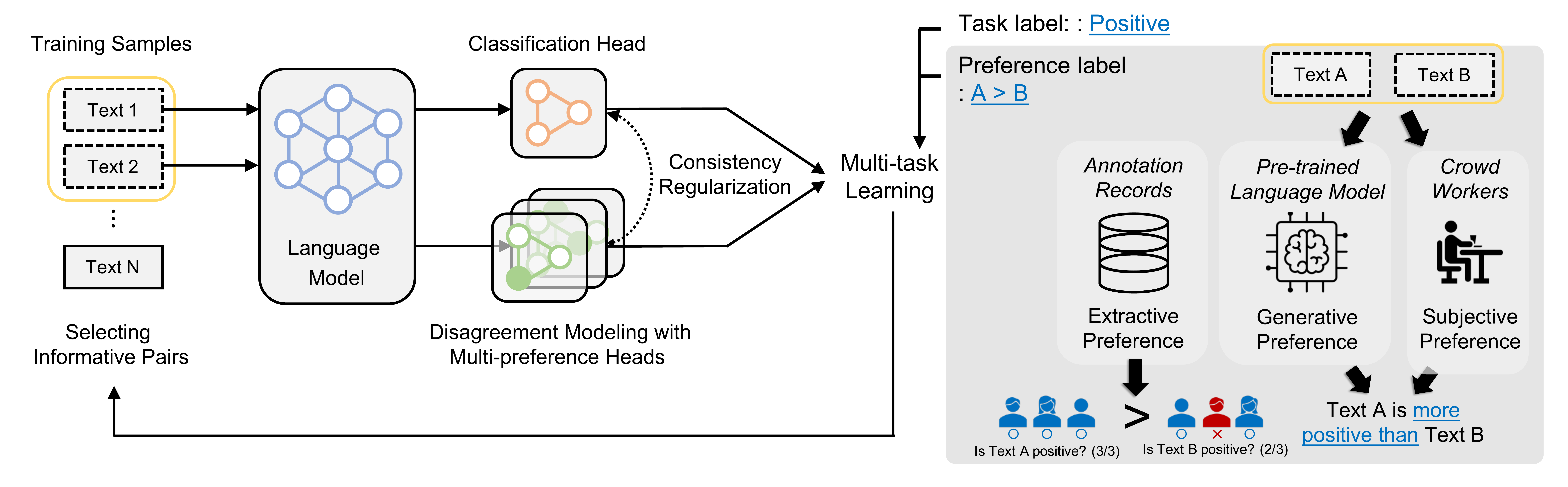}
\end{center}
\vspace{-0.1in}
\caption{
Visual illustration of the proposed auxiliary preference learning for improving the classifier. First, the preference label of the pair of samples is obtained among three different ways (\textit{right}) - \textit{Generative}, \textit{Extractive}, or \textit{Subjective} preference. Then, the preference label is jointly used to train the classifier with the original task label via the proposed \textbf{P}refer-\textbf{to}-\textbf{C}lassify (P2C) framework (\textit{left}).}
\vspace{-0.1in}
\label{fig:illustration}
\end{figure*}

\subsection{Preliminaries}\label{sec:prelimiary}
\textbf{Problem description.} 
We describe the problem setup of our interest under a text classification scenario with $K$ classes. Let $\mathcal{D}$ denote the given training dataset consisting of tuples $(\mathbf{x}, y_{\tt task})\in\mathcal{D}$
where {$\mathbf{x}=[x_{1},\ldots, x_{L}]$ is the sequence of input tokens $x_{i}$, and $y_{\tt task}$ is the target task label.}
Our goal is to train a classifier $f_{\theta}:=W_{\tt task} \circ g_{\phi}$, composed with Transformer-based language model backbone $g_{\phi}$ (\textit{e.g.,} BERT \citep{devlin2019bert}) and a random initialized classification head $W_{\tt task}$, to minimize the task-specific loss $\mathcal{L}_{\tt task}$ such as a cross-entropy loss where $p(\mathbf{x}) = \text{Softmax} \big(f_{\theta}(\mathbf{x})\big)$.

\textbf{Preference learning.} 
In this paper, we use a preference label between two data instances as an auxiliary learning signal to train the classifier.
Specifically, the preference signals reflect the relative suitability between the two input samples with respect to the given task.
We first assume that the preference labels of the given dataset are available. Then, our goal is to train a preference predictor to learn from the given human preferences, by predicting which one among the two input samples is more preferred. To this end, we formulate a preference learning as a supervised learning problem following the approaches in other domains such as reinforcement learning and generative modeling \citep{christiano2017deep, ziegler2019fine, lee2021pebble}.
Given a pair of two different input tokens $(\mathbf{x}^0, \mathbf{x}^1)$ and task label $y_{\tt task}$, a preference label $y_{\tt pref}$ is additionally given; it indicates which input is preferred considering $y_{\tt task}$, \textit{i.e.}, $y_{\tt pref} \in \{0, 1, 0.5\}$, where 1 indicates $\mathbf{x}^1\succ\mathbf{x}^0$ (\textit{i.e.,} $\mathbf{x}^1$ is preferred than $\mathbf{x}^0$), 0 indicates $\mathbf{x}^0\succ\mathbf{x}^1$, and 0.5 implies an equally preferable case.
Each preference label is stored in a dataset $\mathcal{D}$ as a quadruplet $(\mathbf{x}^0, \mathbf{x}^1, y_{\tt task}, y_{\tt pref})$.
Then, we predict a preference using the preference predictor ${h}_\psi$ following \citep{bradley1952rank}:
\begin{align}
    &{\small P_\psi[\mathbf{x}^1\succ\mathbf{x}^0 ; y_{\tt task}] = \frac{\exp \big({h}_{\psi} (\mathbf{x}^1, y_{\tt task})\big)}{\sum_{i\in \{0,1\}} \exp \big({h}_{\psi} (\mathbf{x}^{i}, y_{\tt task})\big)}} \label{eq:pref_model}  
\end{align}
where $\mathbf{x}^i\succ\mathbf{x}^j$ implies that input $i$ is preferable to input $j$. 
The underlying assumption of this model is that the probability of preferring an input depends exponentially on its output.
Then, the preference predictor ${h}_{\psi}$ is trained through supervised learning with the given human preferences, by minimizing the binary cross-entropy loss as follow\footnote{Equally preferable case is learned with the same coefficients.}:

\vspace{-0.2in}
\begin{align}
    \mathcal{L}_{\tt pref}=
    & -\mathop{\mathbb{E}}_{ \substack{(\mathbf{x}^0,\mathbf{x}^1,y_{\tt task} \\ ,y_{\tt pref}) \sim \mathcal{D}}} \Big[y_{\tt pref} \log P_\psi[\mathbf{x}^1\succ\mathbf{x}^0; y_{\tt task}] \nonumber\\
    & + (1-y_{\tt pref})\log P_\psi[\mathbf{x}^0\succ\mathbf{x}^1; y_{\tt task}]\Big] 
    \label{eq:pref_bce} 
\end{align}

\subsection{Prefer-to-classify (P2C)}\label{sec:p2c}

Next, we present the specific techniques to train the classifier with given preference labels: (\textit{a}) multi-task learning of classification and preference learning, (\textit{b}) consistency regularization between classification and preference learning, and (\textit{c}) informative pair sampling method based on the disagreement or inconsistency. 

\textbf{Multi-task learning with preference labels.}
To effectively learn from the given preference label $y_{\tt pref}$ and the task label $y_{\tt task}$, we train the classifier $f_{\theta}$ via multi-task learning \citep{ruder2017overview, sener2018multi} of both classification and preference learning. 
Specifically, we model the preference predictor ${h}_{\psi}$ in Eq. \ref{eq:pref_model} upon the classifier $f_{\theta}$ similar to the case of $W_{\tt task}$. The preference prediction head $W_{\tt pref}$ is added on the output of Transformer backbone $g_{\phi}(\mathbf{x})$ and task label $y_{\tt task}$, \textit{i.e.,} $h_{\psi} (\mathbf{x}, y_{\tt task}) = W_{\tt pref} \circ [g_{\phi}(\mathbf{x}); y_{\tt task}]$\footnote{$[A; B]$ means the concatenation between $A$ and $B$.} where $f_{\theta}(\mathbf{x}) = W_{\tt task} \circ g_{\phi}(\mathbf{x})$.

\textbf{Preference learning with diverse multi-preference heads.} 
In addition, we introduce multiple preference heads $\{W_{\tt pref}^{(t)}\}_{t=1}^{T}$ and trained with $\mathcal{L}_{\tt pref}$ in Eq. \ref{eq:pref_bce} to 
fully exploit the given preference labels.
As obtaining preference labels requires additional cost, it is crucial to find effective ways to exploit them. 
By incorporating multiple preference prediction heads, we can obtain diverse learning signals from each preference label, based on their different random initialization \citep{ganaie2021ensemble}.
However, these multiple preference heads easily collapse into identical ones, as they are built on the compact representation of the pre-trained Transformer shared with the classification head. 
Hence, we introduced diversity regularization between $\{W_{\tt pref}^{(t)}\}_{t=1}^{T}$ during the training; we add a regularization $\mathcal{L}_{\tt div}$ to encourage the diverse prediction for each preference head by maximizing KL-divergence \citep{wang2021long}: 
\begin{align}
    \mathcal{L}_{\tt div} = \frac{-1}{T-1} \sum_{j=1, j \neq i}^{T} D_{\tt KL}\big(
    &P_{\psi^{(i)}}(\mathbf{x}^1, \mathbf{x}^0; y_{\tt task}) || \nonumber \\
    & P_{\psi^{(j)}}(\mathbf{x}^1, \mathbf{x}^0; y_{\tt task}) \big)  \label{eq:diversity}
\end{align}
where {$P_{\psi}(\mathbf{x}^1, \mathbf{x}^0; y_{\tt task})$ is the predictive distribution of the preference predictor ${h}_{\psi}$, \textit{i.e.},} $P_{\psi}(\mathbf{x}^1, \mathbf{x}^0; y_{\tt task}) = \big[P_\psi[\mathbf{x}^1\succ\mathbf{x}^0; y_{\tt task}], P_\psi[\mathbf{x}^0\succ\mathbf{x}^1; y_{\tt task}]\big]$. 
Overall, we train the classifier with the following multi-task learning objective $\mathcal{L_{\tt multi}}$ under hyper-parameter $\lambda_{\tt div}$: 
\begin{equation}
    \mathcal{L}_{\tt multi} = \mathcal{L}_{\tt task} + {\mathcal{L}^{\text{all}}_{\tt pref}} + \lambda_{\tt div} \mathcal{L}_{\tt div} \label{eq:multi_task}
\end{equation}
where $\mathcal{L}^{\psi^{(t)}}_{\tt pref}$ indicate the preference learning objective with each head $\psi^{(t)}$ and $\mathcal{L}^{\text{all}}_{\tt pref} = \sum_{t=1}^{T} \mathcal{L}^{\psi^{(t)}}_{\tt pref}$.

\begin{table*}[t]
\small
\begin{center}
\caption{Examples of the collected generative, extractive, and subjective preference labels on the same pair of sentences.}
\label{tab:examples1}
	\vspace{0.1in}
    \begin{adjustbox}{width=0.9\linewidth}
 \begin{tabularx}{\textwidth}{@{}XX@{}}
		\toprule
    \textbf{A}: I got 3 veggies and a side of fries for over a 11 dollars if you like homecooked food 
    & 
    \textbf{B}: She listened to my ideas, asked questions to get a better idea about my style, and was
excellent at offering advice as if I were a total pleb.
	\\ 
    \multicolumn{2}{@{}c@{}}{ 
    \colorbox{yellow!30}{Sentiment: \underline{Positive}, Generative Preference: \textbf{A $\succ$ B}, Extractive Preference: \textbf{A $\succ$ B}, Subjective Preference: \textbf{B $\succ$ A}}
    }
    \\
    \midrule
    \textbf{A}: We enjoyed our first and last meal in Toronto at Bombay Palace, and I can't think of a better way to book our journey.
    & 
    \textbf{B}: So glad I finally tried this place because if confirmed my suspicions about that critic who rated it a 10.
	\\ 
    \multicolumn{2}{@{}c@{}}{ 
    \colorbox{yellow!30}{Sentiment: \underline{Positive}, Generative Preference: \textbf{A $\succ$ B}, Extractive Preference: \textbf{B $\succ$ A}, Subjective Preference: \textbf{No preference}}
    }
    \\
    \midrule

    \textbf{A}: The buffalo chicken was not good, but very costly.
    & 
    \textbf{B}: There was so much stuff from all over that I had to leave to find an ATM for more cash to pay for it all.
	\\ 
    \multicolumn{2}{@{}c@{}}{ 
    \colorbox{yellow!30}{Sentiment: \underline{Negative}, Generative Preference: \textbf{A $\succ$ B}, Extractive Preference: \textbf{B $\succ$ A}, Subjective Preference: \textbf{B $\succ$ A}}
    }
    \\
    \midrule    

    \textbf{A}: The hotel offered complimentary breakfast.
    & 
    \textbf{B}: My friends had a full acrylic and the other had a fill. It looked so good.
	\\ 
    \multicolumn{2}{@{}c@{}}{ 
    \colorbox{yellow!30}{Sentiment: \underline{Positive}, Generative Preference: \textbf{A $\succ$ B}, Extractive Preference: \textbf{A $\succ$ B}, Subjective Preference: \textbf{A $\succ$ B}}
    }
    \\
    \midrule        

    \end{tabularx}
    \end{adjustbox}
\end{center}
\vspace{-0.1in}
\end{table*}

\textbf{Consistency regularization between classification and preference learning.} 
Even though multi-task learning is an effective way to train the model, it is still unclear whether or not the model can capture the relations between the two tasks explicitly.
Accordingly, we hypothesize that \textit{a more preferred instance should have higher confidence from the classifier}, \textit{i.e.}, $p_{y}(\mathbf{x}^1) > p_{y}(\mathbf{x}^0)$ if $\mathbf{x}^1\succ\mathbf{x}^0$ with the given task label $y$.
Hence, to impose the model explicitly follows this intuition, we further propose a consistency regularization between the two tasks as follows:  
\begin{align}
    \mathcal{L}_{\tt cons} 
    & = y_{\tt pref} \max \{0, p_{y}(\mathbf{x}^1) - p_{y}(\mathbf{x}^0)\} \nonumber\\
    & + (1 - y_{\tt pref}) \max \{0, p_{y}(\mathbf{x}^0) - p_{y}(\mathbf{x}^1)\} \label{eq:consistency_order}
\end{align}
Additionally, when the degree of preference is explicitly provided, \textit{i.e.}, $y_{\tt pref} \in [0,1]$ (see Section \ref{section:extract} of extractive preference case) rather than $y_{\tt pref} \in \{0, 1, 0.5\}$, we further extend this consistency regularization with margin $m$ which represents the degree of preference:  
\begin{align}
    \mathcal{L}_{\tt cons} 
    & = y_{\tt pref} \max\{0, m - \Delta p_{y}(\mathbf{x}^1, \mathbf{x}^0)\} \nonumber\\ & + (1 - y_{\tt pref}) \max \{0, \Delta p_{y}(\mathbf{x}^1, \mathbf{x}^0) - m\} \label{eq:consistency_gap}
\end{align}
where $\Delta p_{y}(\mathbf{x}^1, \mathbf{x}^0) = p_{y}(\mathbf{x}^0) - p_{y}(\mathbf{x}^1)$. We note that the previous consistency regularization Eq. \ref{eq:consistency_order} becomes the special case of Eq. \ref{eq:consistency_gap} with $m=0$.
Overall, our training loss of the classifier is as follows:
\begin{equation}
    \mathcal{L}_{\tt train} = \mathcal{L}_{\tt multi} + \lambda_{\tt cons} \mathcal{L}_{\tt cons} \label{eq:overall}
\end{equation}
where $\lambda_{\tt cons}$ is a hyper-parameter.

\textbf{Selecting informative pairs.}
As the number of pairs of samples $(\mathbf{x}^0, \mathbf{x}^1)$ is proportional to the square of the number of training samples, it is difficult to obtain the preference label for all possible pairs, and even harder to learn from them even if we have all the preference labels.
Hence, we propose the following advanced sampling scheme to maximize preference learning's effectiveness during training: (1) \textit{Disagreement-based} sampling, which selects pairs of instances with high variance across multiple preference predictors $\{{h}_{\psi^{(i)}}\}_{i=1}^{T}$, and (2) \textit{Inconsistency-based} sampling, which seeks to reduce the mismatched pairs with high consistency loss $\mathcal{L_{\tt cons}}$ in Eq. \ref{eq:consistency_order}.
We evaluate the effects of these sampling methods in Appendix \ref{sup:ablation}.
\section{Collection of Preference Labels}\label{sec:preference_collection}

In this section, we present the descriptions of three different types of preference labels (\textit{generative}, \textit{extractive}, and \textit{subjective}) to apply auxiliary preference learning via \name{}.
The detailed procedure of collecting each preference label and comparison between them is presented in Appendix \ref{sup:interface}.

\textbf{Generative preference from large language models.}
First, we propose to use the recent generative pre-trained large language models \cite{brown2020language, ouyang2022training} to obtain the preference between pair of samples, and call the obtained preference label as \textit{generative preference}. 
These models have recently demonstrated the strong zero/few-shot generalization performance in various NLP tasks, and our high-level intuition is that such capability could be effective to provide a useful signal between the samples. 
To be specific, we use GPT-3 through the officially provided API,\footnote{\texttt{text-davinci-003} in \url{https://beta.openai.com/docs/models/gpt-3}} by querying the pair of sentences along with the properly designed prompts, presented in Appendix \ref{sup:interface}. 
For the experiments of the dataset with $N$ samples, we randomly select one pair for each sample and acquire $N$ generative preference labels for \name{}. 

\textbf{Extractive preference from data annotation records.}
Another way is to extract the preference signals from the existing datasets; our high-level assumption is that annotation records of each data, which are naturally gathered during the construction of the dataset, implicitly capture the preference between data samples.
For example, if one sample has higher voting (9 out of 10) than the other sample (6 out of 10) as positive sentiment, one can assume that the former has a relatively higher preference. 
We call this implicit preference label as \textit{extractive preference}; since the extractive preference is derived from existing sources of the dataset, it can be obtained for any pair of samples \textit{without additional cost}. 
Hence, for the dataset with $N$ samples, one can obtain $N^2$ of extract preference labels at maximum. 
We randomly sample the pair of each sample and use their preference labels during training for \name{}.

\textbf{Subjective preference from crowd workers.} 
Lastly, we consider directly collecting the human preference and call it as \textit{subjective preference}; while it requires a high payment to hire human annotators, it is expected to be the most accurate as it is directly obtained by asking humans. 
Hence, to investigate the advantage of human preference, we construct the subjective preference dataset based on DynaSent-R2 dataset \citep{potts2021dynasent} for the sentiment classification task. 
Specifically, we gather the subjective preference of the sentence pairs by asking crowd workers to answer “\textit{which sentence is more positive (neutral, or negative)?}” using Amazon's Mechanical Turk crowd-sourcing platform \citep{crowston2012amazon}. 
Then, each worker should select one of the two sentences or answer “No Preference”. 
Following \citep{nie2020adversarial}, we hire three crowd workers for each pair of sentences at the most, and the pairs are dynamically selected across multiple rounds to maximize the obtained information.
Consequently, we collect a total of 5,000 pairs' subjective preference labels. 
 
As described above, each type of preference has distinct characteristics as shown in Table \ref{table:comparison}; extractive preference could be freely obtained if the annotation records of the benchmark are available (\textit{i.e.}, lowest cost). 
On the other hand, generative preference may require an additional cost, but it is not expensive and provides the easiest way to access the preference labels.
While subjective preference is the most expensive (e.g., 1.6\$ for 10 samples, while 8.0\$ for 5,000 samples with GPT-3), it has a clear advantage of providing an accurate and human-aligned preference signal.
To verify the effect of those differences, we present qualitative and quantitative examples in Table \ref{tab:examples1} and Appendix \ref{sup:interface}.

\begin{table}[t]
\vspace{-0.1in}
\centering
\caption{Comparison of three different types of preference labels.}
\vspace{0.1in}
\scalebox{0.9}{
\begin{tabular}{r|ccc}
\toprule
\textbf{Types} & \textbf{Cost} & \textbf{Accuracy} & \textbf{Accessibility} \\ \midrule
Generative Preference & Medium & Medium & {\color{blue} High} \\ 
Extractive Preference & {\color{blue} Low} & Medium & Medium \\ 
Subjective Preference & {\color{red} High} & {\color{blue} High} & {\color{red} Low} \\ \bottomrule
\end{tabular}
}
\label{table:comparison}
\vspace{-0.2in}
\end{table}
\section{Experiments}\label{section:experiments}

\begin{table*}[t]
	\begin{center}
	\caption{Test accuracy of fine-tuned RoBERTa classifiers with specified training methods or GPT-3 with few-shot prompting on four different text classification datasets. For \name{}, the generative preference labels are obtained from GPT-3. All the values and error bars are mean and standard deviation across 5 random seeds. The best and the second best results are indicated in \textbf{bold} and \underline{underline}, respectively. In the case of few-shot GPT-3, we obtain standard deviation by 3 runs over randomly sampled few-shot examples for prompting.}
	\vspace{0.0in}
	\label{table:gpt_only}
    \scalebox{0.85}{
	\begin{tabular}{r|cc|cc|cc|cc}
		\toprule
		& \multicolumn{2}{c}{CoLA} &
		\multicolumn{2}{c}{SMS Spam} &
		\multicolumn{2}{c}{Hate Speech} &\multicolumn{2}{c}{Emotion} \\
		Method  & $\text{Mcc} (\uparrow)$ 
		& $\text{ECE} (\downarrow)$ 
		& $\text{bAcc} (\uparrow) $ / wAcc$(\uparrow)$ 
		& $\text{ECE} (\downarrow)$ 
		& bAcc$(\uparrow)$ / wAcc$(\uparrow)$ 
		& $\text{ECE} (\downarrow)$ 
		& bAcc$(\uparrow)$ / wAcc$(\uparrow)$ 
		& $\text{ECE} (\downarrow)$ \\ \midrule
		
		Vanilla       & {63.7\ms{1.0}}
		              & \underline{3.6}\ms{1.6}
		              & {96.9\ms{0.3}} / {\underline{95.1}\ms{1.5}}
		              & {1.3\ms{0.3}}
		              & {81.1\ms{1.8}} / {69.9\ms{4.6}}
		              & {5.1\ms{1.0}}
		              & {88.6\ms{2.3}} / {76.1\ms{7.8}}
		              & {4.0\ms{1.1}} \\ 
		Label Smoothing  & {63.9\ms{0.3}}
		              & {4.6\ms{1.2}}
		              & {96.9\ms{0.8}} / {94.0\ms{1.5}}
		              & {1.1\ms{0.3}}
		              & {81.5\ms{0.9}} / {\underline{71.3}\ms{3.2}}
		              & {6.6\ms{1.0}} 
		              & {\underline{89.8}\ms{0.8}} / {\underline{76.9}\ms{6.6}}
		              & {4.0\ms{0.9}}\\ 
		Max Entropy  & {64.1\ms{0.3}}
		              & {4.5\ms{0.4}}
		              & {\underline{96.9}\ms{1.1}} / {94.7\ms{1.6}}
		              & {1.2\ms{0.3}}
		              & {\underline{81.6}\ms{1.8}} / {70.5\ms{4.2}}
		              & \underline{4.3}\ms{0.7}
		              & {89.1\ms{1.1}} / {73.1\ms{2.5}}
		              & \textbf{3.6}\ms{0.9}
		              \\ 
		CS-KD         & \underline{64.5}\ms{1.4}
		              & {4.1\ms{1.1}}
		              & {96.8\ms{0.9}} / {94.0\ms{2.4}}
		              & \textbf{1.1}\ms{0.2}
		              & {81.4\ms{2.6}} / {69.6\ms{5.1}}
		              & {5.3\ms{1.8}} 
		              & {89.4\ms{1.6}} / {74.0\ms{6.8}}
		              & {{4.1}\ms{0.2}}\\ 
		GPT-3 (0-shot)& {60.4}
		              & - 
		              & {90.3} / {84.3}
		              & -
		              & {68.7} / {41.6}
		              & -
		              & {50.2} / {23.3}
		              & - \\ 
	    GPT-3 (5-shot)& {58.5}\ms{0.4}
		              & -
		              & {{92.2}\ms{0.5}} / {88.5\ms{0.7}}
		              & -
		              & {{78.5}\ms{2.0}} / {{70.3}\ms{3.6}}
		              & -
		              & {{46.6}\ms{0.6}} / {{30.3}\ms{2.6}}
		              & - \\ 
		GPT-3 (20-shot)& {58.3}\ms{1.4}
		              & -
		              & {{95.8}\ms{0.4}} / {94.4\ms{0.7}}
		              & -
		              & {{77.8}\ms{0.5}} / {{69.0}\ms{1.5}}
		              & -
		              & {{47.5}\ms{1.0}} / {{30.8}\ms{4.5}}
		              & - \\ 
		              \midrule
		P2C (Ours)    & \textbf{65.4}\ms{1.0}
		              & \textbf{2.8}\ms{1.1}
		              & {{\textbf{97.4}}\ms{0.4}} / {\textbf{95.2}\ms{1.0}}
		              & {\textbf{1.1}}\ms{0.3}
		              & {\textbf{82.4}\ms{1.3}} / {\textbf{73.6}\ms{4.5}}
		              & \textbf{4.0}\ms{0.3}
		              & {\textbf{90.7}\ms{0.7}} / {\textbf{81.7}\ms{4.7}}
		              & \textbf{3.6}\ms{0.8} \\ \bottomrule
	\end{tabular}
    }
    \end{center}
    \vspace{-0.1in}
\end{table*}

\subsection{Setups}\label{section:setups}

\textbf{Datasets.} 
For the experiments, we first use the following four text classification datasets: (1) \textit{CoLA} \cite{warstadt2019neural}, (2) \textit{SMS Spam} \cite{almeida2011contributions}, (3) \textit{Hate Speech} \cite{fivser2018proceedings}, and (4) \textit{Emotion} \cite{saravia2018carer}.  
In addition, to demonstrate the effectiveness of extractive preference, we investigate the publicly available datasets providing the annotation records and use the following six text classification datasets.
DynaSent \citep{potts2021dynasent} is a dynamically constructed sentiment classification benchmark with ternary (positive/negative/neutral) sentiments; we use the dataset from the first round, (5) \textit{DynaSent-R1}, and from the second round, (6) \textit{DynaSent-R2}, the dataset for our experiments. 
Standford politeness corpus \citep{danescu2013polite} is a binary classification benchmark for predicting whether the given sentence is polite or impolite. 
Since there are two different input domains within this benchmark, we split them into two different datasets: (7) \textit{Polite-Wiki} from Wikipedia, and (8) \textit{Polite-SE} from StackExchange, following the original setup.
(9) \textit{Offensive} agreement dataset \citep{leonardelli2021agreeing} is a binary classification benchmark for predicting whether the given sentence is offensive or not.
(10) \textit{MultiNLI} \citep{williams2018broad} is a crowd-sourced collection of sentence pairs annotated with textual entailment information; as the only validation set includes the annotation records, we split it into 8:1:1 for training, validation, and test sets.
All datasets have the annotation records from 5 annotators for each sample.
More details of datasets are presented in Appendix \ref{sup:datasets}.

\textbf{Baselines.} 
We first compare the proposed P2C to a na\"ive training with a cross-entropy loss and majority voted task label, denoted by (a) \textit{Vanilla}. 
Then, since \name{} with extractive preference can be viewed as a new way to utilize the annotation records, we compare this method with a wide range of disagreement learning methods in Section \ref{section:extract}, as listed as follows;
(b) \textit{Soft-labeling} \citep{fornaciari2021beyond}: using the probabilistic distribution of annotations as soft labels for training;
(c) \textit{Margin} \citep{sharmanska2016ambiguity}: training the model with hinge loss by setting a margin proportional to the annotators' agreements;
(d) \textit{Filtering} \citep{leonardelli2021agreeing}: removing the training samples with a high disagreement. 
(e) \textit{Weighting} \citep{uma2021learning}: using weighted cross-entropy with smaller weights for the samples with high disagreements; 
(f) \textit{Multi-annotator} \citep{davani2022dealing}: training the multiple classification heads for each annotation and using its ensemble for the evaluation. 
Furthermore, since we train the model with pair of samples, we also consider the baseline considering pair-wise training, 
(g) Class-wise Self-Knowledge Distillation (\textit{CS-KD}) \citep{yun2020regularizing}, which forces the similar predictive distribution between the same class samples to be similar. 
Lastly, we consider two regularization methods, (h) \textit{Label Smoothing} \cite{muller2019does} and (i) \textit{Max Entropy} \cite{pereyra2017regularizing}, as the baselines of \name{} with generative preference in Section \ref{sec:generative}. 
Details are described in Appendix \ref{sup:baselines}.

\textbf{Implementation details.} 
All the experiments are conducted by fine-tuning RoBERTa-base \citep{liu2019roberta} using Adam optimizer \citep{kingma2014adam} with a fixed learning rate 1e-5 and the default hyper-parameters of Adam. 
For all datasets, the model is fine-tuned using the specified training method with batch size 16 for 20 epochs.
In the case of P2C, we use $T=3$ preference heads $\{W_{\tt pref}^{(i)}\}_{i=1}^{T}$ and 2-layer MLPs for each $W_{\tt pref}$. 
We choose hyper-parameters from a fixed set of candidates based on the validation set: $\lambda_{\tt cons},\lambda_{\tt div}  \in \{1.0, 0.1\}$.
We sample the pair of instances with the same task labels for efficiency.
With the extractive preference, we apply the consistency loss with margin (Eq. \ref{eq:consistency_gap}) by using the difference of annotation as the margin $m$. For other cases, we apply \name{} with consistency loss without margin (Eq. \ref{eq:consistency_order}) on the pre-defined pairs of samples.
More details and experimental supports for the design choices can be found in Appendix \ref{sup:p2c} and \ref{sup:ablation}, respectively.

\subsection{Experiments with generative preference}\label{sec:generative}

In this section, we first evaluate our framework with the generative preference labels obtained from the pre-trained large language model, GPT-3 \cite{brown2020language}.
To validate the effectiveness of \name{} under the more challenging scenario, we use the following four datasets which have a skewed label distribution without annotation records: \textit{CoLA}, \textit{SMS Spam}, \textit{Hate Speech}, and \textit{Emotion}. 
Since their test datasets are also imbalanced, we measure the balanced accuracy (\textit{bAcc}) \cite{huang2016learning} and the worst-group accuracy (\textit{wAcc}) \cite{sagawa2020distributionally}, to evaluate the generalization capability of the model, except CoLA since it is usually used with own metric, Matthews correlation coefficient (\textit{Mcc}) \cite{chicco2020advantages}.
In addition, to measure the calibration of the trained model, we report Expected Calibration Error (\textit{ECE}) \cite{guo2017calibration}.
Here, we commonly adopt the temperature scaling to measure ECE following \citep{guo2017calibration}. 
As the annotation records are unavailable, we compare \name{} with the baseline methods incurring the smoothed prediction of classifier only using the task label: \textit{Label Smoothing}, \textit{Entropy Maximization}, and \textit{CS-KD}. 
In addition, we use $K$-shot prompting predictions of GPT-3 ($K=0, 5, 20$) as an additional baseline. 

As shown in Table \ref{table:gpt_only}, the generative preference labels with \name{} are consistently effective in improving the performance of the text classifier; for example, \name{} exhibited 11.55\% relative test error reduction on average compared to \textit{Vanilla} while also improving the predictive calibration. 
At the same time, we note that \name{} shows better performance than the considered baseline, which indicates that the training signal from the preference label is more than smoothing the prediction of the classifier.
Finally, as shown in Table \ref{table:gpt_only}, P2C significantly outperforms GPT-3 baselines, which means that our framework does not just distill the `instance-wise' knowledge of GPT-3, but obtains complementary information through the proposed `pair-wise' comparisons.

\begin{table*}[t]
	\begin{center}
	\caption{Test accuracy of fine-tuned RoBERTa classifiers with each annotation method on six different text classification datasets. For \name{}, the extractive preference labels are obtained from the annotation records of each dataset. All the values and error bars are mean and standard deviation across five random seeds. The best and the second best results are indicated in \textbf{bold} and \underline{underline}, respectively.}
	\label{table:main_extract}
	\vspace{0.1in}
    \scalebox{0.9}{
	\begin{tabular}{r|cccccc}
		\toprule
		Method & Offensive & Polite-Wiki & Polite-SE 
		& MNLI & DynaSent-R1 & DynaSent-R2   
		
		              \\ \midrule
		Vanilla       & {75.88\ms{0.72}} 
		              & {89.35\ms{1.53}}
		              & {70.00\ms{1.49}}  
		              & {81.92\ms{0.70}} 
		              & {80.43\ms{0.30}}
		              & {71.23\ms{1.05}}
		               \\
		Soft-labeling & {76.08\ms{1.44}}
		              & {89.57\ms{1.76}}
		              & {70.35\ms{1.68}}  
		              & \underline{82.67}\ms{0.50}
		              & {81.10\ms{0.33}}
		              & \underline{72.15}\ms{1.59}
		               \\
		Margin Loss   & \underline{76.67}\ms{1.18}
		              & {88.51\ms{0.93}}
		              & \underline{70.51}\ms{1.16} 
		              & {81.41\ms{0.63}} 
		              & {80.42\ms{0.23}}
		              & {69.27\ms{0.98}}
		               \\
		Filtering     & {76.13\ms{1.18}} 
		              & {89.50\ms{0.87}}  
		              & {68.28\ms{2.43}}
		              & {82.13\ms{0.67}} 
		              & {80.38\ms{0.34}}
		              & {69.86\ms{0.78}} 
		               \\
		Weighting    & {76.17\ms{1.18}} 
		              & {89.65\ms{1.46}}  
		              & {68.38\ms{1.67}}
		              & {82.48\ms{0.49}} 
		              & {80.21\ms{0.41}}
		              & {71.81\ms{1.12}}
		               \\
		Multi-annotator  & {76.50\ms{1.98}} 
		              & \underline{89.88}\ms{1.82}
		              & {69.39\ms{2.84}}  
		              & {82.61\ms{0.70}} 
		              & \underline{81.14}\ms{0.55}
		              & {71.97\ms{1.25}}
		               \\
		CS-KD         & {75.75\ms{0.66}} 
		              & {89.65\ms{1.84}}
		              & {70.10\ms{1.29}}  
		              & {82.32\ms{0.23}} 
		              & {80.63\ms{0.27}}
		              & {71.81\ms{0.67}}
		               \\    \midrule
		P2C (Ours)    & \textbf{77.81\ms{0.21}} 
		              & \textbf{91.06\ms{0.64}}
		              & \textbf{71.21\ms{0.93}}  
		              & \textbf{83.15\ms{0.29}} 
		              & \textbf{81.50\ms{0.39}}
		              & \textbf{73.06\ms{0.31}}
		          
		               \\\bottomrule
	\end{tabular}
    }
    \end{center}
    \vspace{-0.1in}
\end{table*}
\begin{figure*}[t]
\begin{center}
    {
    \subfigure[Disagreement Prediction ($\downarrow$)]
        {
        \includegraphics[width=0.31\textwidth]{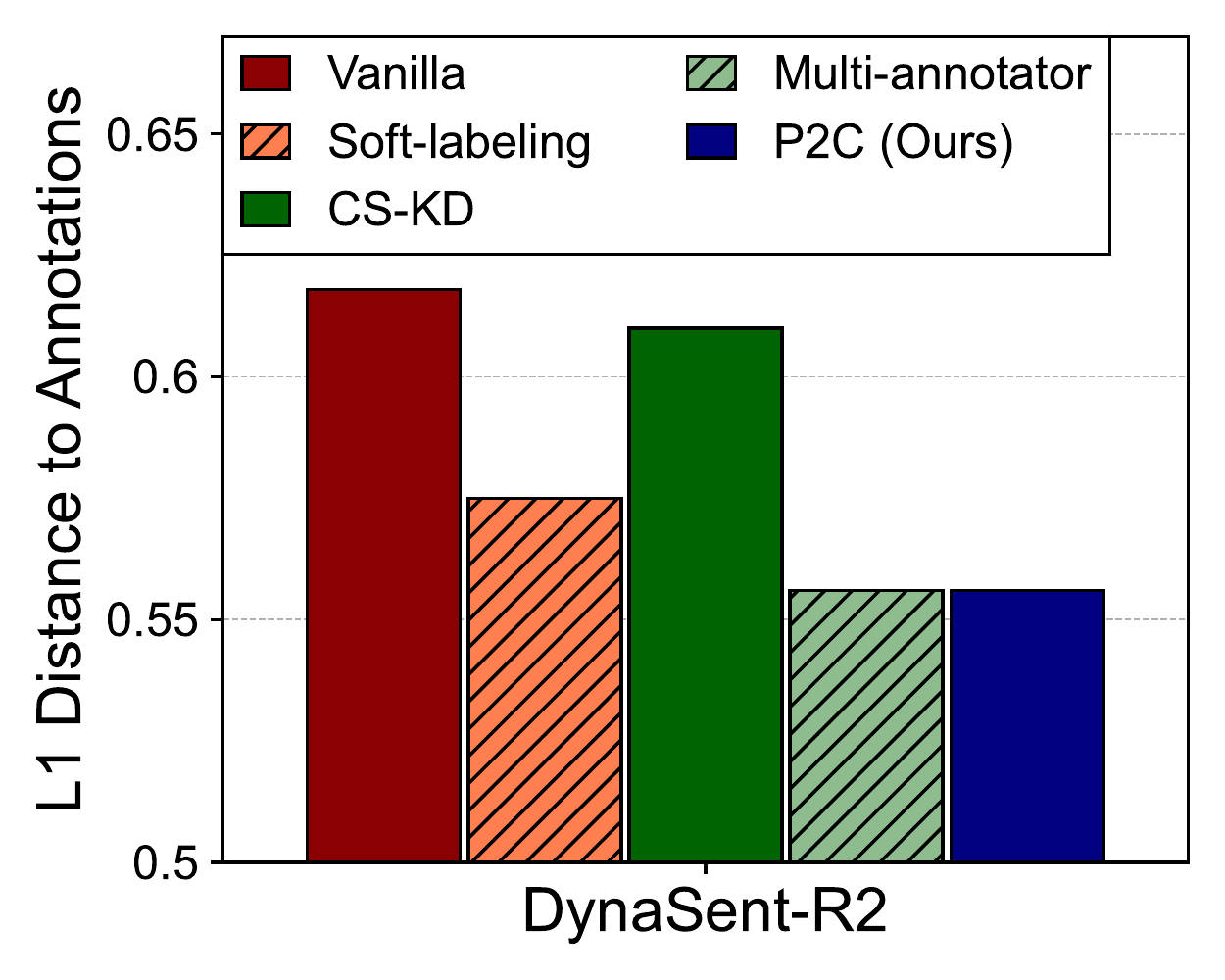}
        \label{fig:2a}
        }
    \subfigure[Reliability Diagram ($\diagup$)]
        {
        \includegraphics[width=0.31\textwidth]{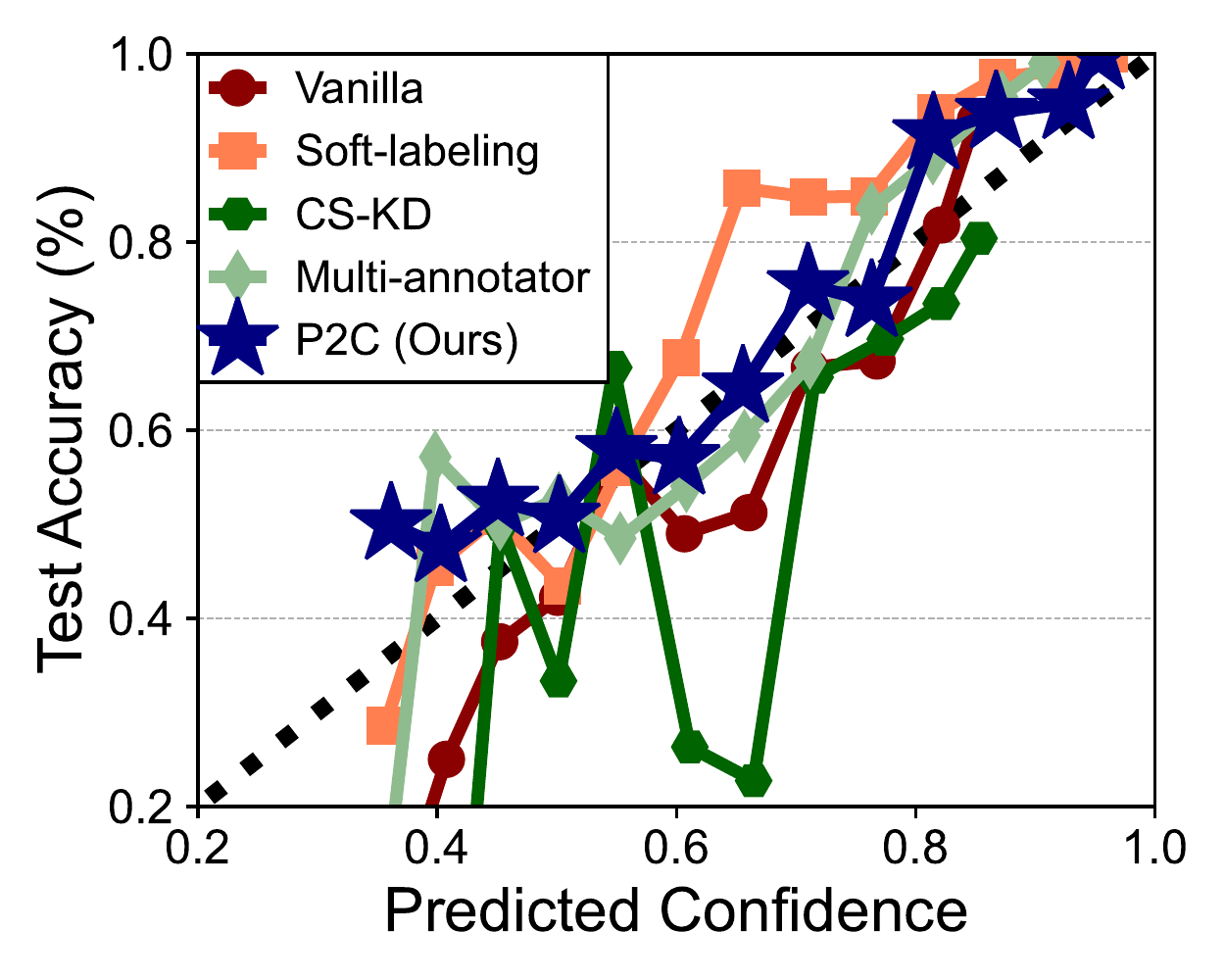}
        \label{fig:2b}
        }
    \subfigure[Expected Calibration Error ($\downarrow$)]
        {
        \includegraphics[width=0.31\textwidth]{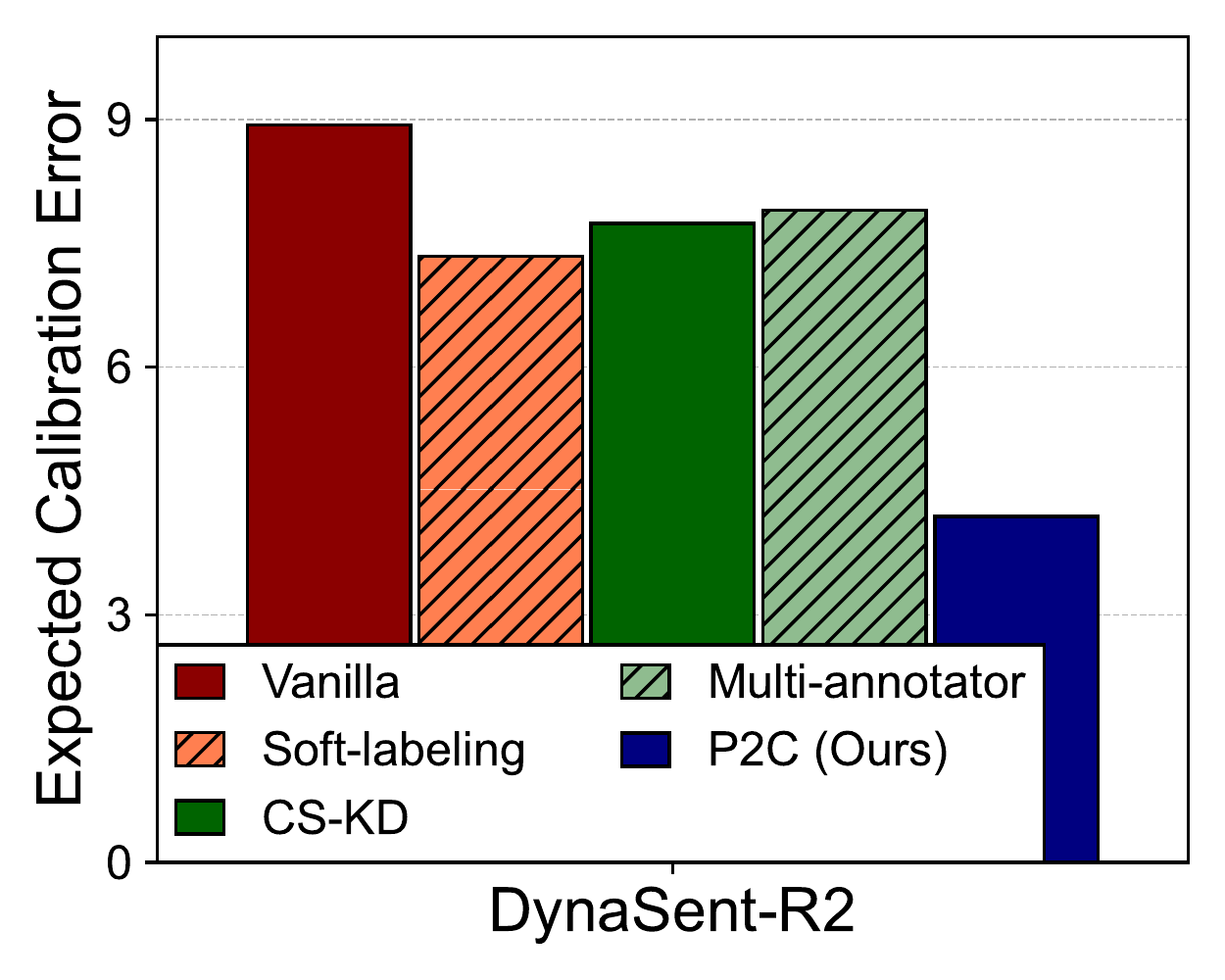}
        \label{fig:2c}
        } 
    }
\end{center}
\vspace{-0.1in}
\caption{Additional experimental results of P2C with extractive preference on DynaSent-R2. 
(a) Average L1 distance between the predictions and the soft labels obtained from the annotation records. The lower distance ($\downarrow$) means better alignment with annotators.
(b) Reliability diagram shows accuracy as a function of confidence. Perfect calibration is plotted by dashed diagonals ($\diagup$). 
(c) Expected Calibration Error (ECE) to quantitatively measure the calibration of the classifier. 
The lower ECE ($\downarrow$) means better calibration.
}
\vspace{-0.1in}
\label{fig:analysis}
\end{figure*}

\subsection{Experiments with extractive preference}\label{section:extract}

\begin{table*}[t]
	\begin{center}
	\caption{Ablation study with each component of P2C with extractive preference labels. Test accuracy of fine-tuned RoBERTa classifiers on DynaSent-R2 and Offensive are compared. All the values and error bars are mean and standard deviation across 5 random seeds.}
	\vspace{0.1in}
	\label{table:ablation}
    \scalebox{0.9}{
	\begin{tabular}{r|cccccc|cc}
		\toprule
		Method & $T$ & $\mathcal{L}_{\tt task}$ & $\mathcal{L}_{\tt pref}$  & $\mathcal{L}_{\tt div}$ & $\mathcal{L}_{\tt cons}$ & Sampling & DynaSent-R2 & Offensive 
		              \\ \midrule
		Vanilla       & -  
		              & \checkmark
		              & -
		              & -
		              & -
		              & -
		              & {71.23\ms{1.05}}  
		              & {75.88\ms{0.72}} 
		               \\ \midrule
		Preference & 1 & \checkmark
		              & \checkmark
		              & -
		              & -
		              & -
		              & {71.84\ms{0.78}}  
		              & {75.90\ms{1.15}} 
		               \\
		              & 3 & \checkmark
		              & \checkmark
		              & -
		              & -
		              & -
		              & {71.92\ms{0.66}} 
		              & {76.43\ms{0.32}} 
		               \\
		              & 3 & \checkmark
		              & \checkmark
		              & \checkmark
		              & -
		              & -
		              & {72.05\ms{1.30}}  
		              & {76.67\ms{1.38}} 
		              \\ 
		              & 3 & \checkmark
		              & \checkmark
		              & \checkmark
		              & \checkmark
		              & -
		              & {72.67\ms{0.89}}  
		              & {77.67\ms{0.99}} 
		               \\ \midrule
		P2C (Ours)    & 3      & \checkmark
		              & \checkmark
		              & \checkmark
		              & \checkmark
		              & \checkmark
		              & {73.06}\ms{0.31}  
		              & {77.81}\ms{0.21} 
		               \\\bottomrule
	\end{tabular}
    }
    \end{center}
    \vspace{-0.15in}
\end{table*}

\begin{table*}[t]
	\begin{center}
	\caption{Results of fine-tuned RoBERTa classifiers with 
	different ways to obtain the labels on DynaSent-R2. $N_{\tt task}$ and $N_{\tt pref}$ indicate the number of used task labels and preference labels, respectively. $\text{d}_{\tt hard}$ and $\text{d}_{\tt easy}$ are the $l_{1}$ distance to annotations with hard and easy samples. Here, the difficulty is defined on the disagreement between annotators. All the values and error bars are mean and standard deviation across 5 random seeds. The best and the second best results are indicated in \textbf{bold} and \underline{underline}, respectively.}
	\vspace{0.1in}
	\label{table:subject}
    \scalebox{0.9}{
	\begin{tabular}{r|cc|ccccc}
		\toprule
		Method  & $N_{\tt task}$ & $N_{\tt pref}$ & $\text{Acc}_{\tt avg} (\uparrow)$ 
		& $\text{Acc}_{\tt hard}$ / $\text{Acc}_{\tt easy} (\uparrow)$  
		& $\text{ECE} (\downarrow)$ 
		& $\text{d}_{\tt hard}$ / $\text{d}_{\tt easy} (\downarrow)$ \\ \midrule
		{Vanilla}      
		              & 7.5k & -
		              & {69.03\ms{1.29}}
		              
		              & {59.33\ms{2.57}} / {80.00}\ms{1.22}
		              & {9.25\ms{1.39}}
		              
		              & {0.856\ms{0.01}} / {0.405\ms{0.03}}
		              
		               \\ \midrule
		Task Labels   & 12.5k & - 
		              & {71.17\ms{1.35}}
		              
		              & {57.86\ms{2.31}} / \textbf{84.21}\ms{1.05}
		              & {9.19\ms{1.36}}
		              
		              & {0.878\ms{0.04}} / {\textbf{0.327}\ms{0.02}}
		               \\
		Generative Preference & 7.5k & 5k
		              & \underline{71.46}\ms{1.16}
		              & {\underline{61.77}\ms{0.94}} / {82.28\ms{1.01}} 
		              & \underline{6.64}\ms{0.79}
		              & {0.850\ms{0.02}} / {0.361\ms{0.02}} 
		              
		               \\
		Extractive Preference & 7.5k & 5k
		              & {71.36\ms{1.19}}
		              & {61.16\ms{1.91}} / \underline{83.11}\ms{1.78}
		              & {6.75\ms{0.78}}
		              & {\underline{0.847}\ms{0.03}} / {\underline{0.351}\ms{0.03}} 
		              
		               \\
		Subjective Preference & 7.5k & 5k
		              & \textbf{71.74}\ms{1.04}
		              
		              & {\textbf{62.08}\ms{0.94}} / {83.01}\ms{1.27}
		              & \textbf{6.09}\ms{0.31}
		          
		              & {\textbf{0.828}\ms{0.02}} / {0.356\ms{0.02}}
		              
		              \\ \bottomrule
	\end{tabular}
    }
    \end{center}
    \vspace{-0.15in}
\end{table*}
While the generative preference is an efficient way to apply auxiliary preference learning with \name{} using the moderate cost, it would be much better if one could still benefit from \name{} for free. 
In this section, we evaluate the effectiveness of \name{} with the extractive preference labels, which could be freely obtained from the existing benchmarks if the annotations records are available.
We compare P2C with various disagreement learning schemes to fine-tune the RoBERTa-base classifier for each dataset, as they also utilize the annotation records for better training. 
Table \ref{table:main_extract} summarizes the results on six text classification datasets. 
Remarkably, P2C consistently outperforms the baseline methods for all six datasets. 
To be specific, P2C exhibits 7.59\% relative test error reduction compared to the vanilla method in the average. 
Furthermore, compared to the previous best disagreement learning method of each dataset, P2C exhibits 4.27\% relative test error reduction on average. 
These results show that extractive preferences successfully provide complementary training signals to the classifier from the pair-wise preference, and demonstrate the effectiveness of P2C as a training method to utilize the annotation records.

Next, on DynaSent-R2, we conduct additional experiments to verify how P2C improves the classifier. 
We first check whether the prediction of the trained model with P2C is similar to the annotators' judgment, as the extractive preference labels come from annotation records. 
Specifically, we compare the L1 distance between the predictions of the model and the soft labels from the annotation records in Figure \ref{fig:2a}. 
We verify that P2C achieves the lowest distance to the soft labels, showing the validity of our preference learning for better modeling of the given task. 
Moreover, we verify that the calibration of the classifier is more improved than the baselines, as a result of pair-wise preference modeling.
To be specific, we provide a reliability diagram \citep{yun2020regularizing}, which plots the expected sample accuracy as a function of the confidence of the classifier in Figure \ref{fig:2b}.
We remark that the plotted identity function (dashed diagonal) implies perfect calibration \citep{guo2017calibration}, and our method is the closest one among the baselines.
This calibration effect of P2C is further verified through ECE in Figure \ref{fig:2c}. 

To validate the effectiveness of the proposed component of P2C in Section \ref{sec:p2c}, we perform the ablation experiments, and the results are presented in Table \ref{table:ablation}, as the extractive preference of all pairs is accessible. 
It is observable that diverse multi-preference heads improve the effectiveness of preference labels with better modeling compared to the single preference head (\textit{2-4th rows}).
In addition, consistency regularization between classification and preference heads enables the classifier to fully utilize the pair-wise training signal to solve the task, hence the performance is significantly improved (\textit{5th row}). 
The performance is further improved by selecting the informative pairs during the training (\textit{6th row}). More results are in Appendix \ref{sup:ablation}.

\subsection{Experiments with subjective preference}\label{section:experiments_subject}

Lastly, we verify the effectiveness of collected subjective preference labels compared to other types of labels. 
To this end, we consider the scenario that the specific types of labels are additionally obtained on top of the existing task labels.
Namely, task labels could be collected more with additional training samples, or preference labels between the existing samples could be obtained.
Table \ref{table:subject} summarizes the experimental results. 
Here, it is observed that subjective preference labels are the most effective for improving the test accuracy ($\text{Acc}_{\tt avg}$) along with a better calibration effect.
Remarkably, it is noticeable that the preference labels significantly improve the accuracy on relatively hard samples ($\text{Acc}_{\tt hard}$) regardless of its type, while the additional task labels are effective for the relatively easy samples.\footnote{We define the difficulty based on the disagreement of annotators, \textit{i.e.,} more disagree indicates more difficult.}
Somewhat surprisingly, one can observe that the additional 5k generative preference labels by GPT-3 are more effective than the same number of task labels, although the former is much cheaper to obtain than the latter; it indicates that our framework can serve as a new effective way to evolve the existing benchmarks along with the recent development of pre-trained large language models at a considerable cost.  

\subsection{Applications of P2C beyond text classification}\label{section:exp.vision}

\begin{table*}[t]
    \begin{center}
	\caption{Test accuracy of ResNet-18 classifiers on five different binary attribute classification tasks, following the setups in \citep{sharmanska2016ambiguity}. For \name{}, the extractive preference labels are obtained from the annotation records of each attribute. All the values and error bars are mean and standard deviation across five random seeds. The best results are indicated in \textbf{bold}.}
	\vspace{0.1in}
	\label{table:supp_vision}
        \scalebox{0.9}{
	\begin{tabular}{r|ccccc|c}
		\toprule
		Method / Tasks & nat.light & man-made & open & enclosed & nohorizon & Average \\ \midrule
		Vanilla      
		                & {78.67\ms{1.05}} 
                            & {62.50\ms{2.65}} 
		                & {79.00\ms{1.43}} 
		                & {85.83\ms{0.61}}  
		                & {76.42\ms{0.62}}
		                & {76.48\ms{1.25}}
		               \\ 
		P2C (Ours)          
		              & \textbf{79.83}\ms{0.94}
                        & \textbf{64.83}\ms{1.71}
		              & \textbf{82.33}\ms{0.72}   
        		      & \textbf{86.25}\ms{1.43}  
                        & \textbf{77.00}\ms{1.06}    
        		    & \textbf{78.05}\ms{1.17}    
		               \\\bottomrule
	\end{tabular}
        }
    \end{center}
    \vspace{-0.15in}
\end{table*}

While we primarily demonstrate the effectiveness of \name{} on the text classification datasets, our approach has the potential to be applicable beyond text classification.
To empirically verify such advantages, we have conducted additional experiments on an image classification task to validate our approach's applicability. 
Specifically, we used the publicly available SUN Attribute dataset \citep{patterson2014sun} and constructed multiple binary scene attribute classification tasks from it, following the setups in \citep{sharmanska2016ambiguity}.
Here, we considered the largest five attributes for the experiments.
As this dataset includes annotation records, we constructed extractive preference labels to apply \name{}. 
For experiments, we commonly trained ResNet-18 \citep{he2016deep} from scratch, for 100 epochs using the SGD optimizer with a weight decay of 0.01 and a learning rate of 0.1 (decreased to 0.01 and 0.001 at 50 and 75 epochs, respectively).

In Table \ref{table:supp_vision}, one can observe that our P2C approach effectively improves the performance of the image classifier, with an average relative test error reduction of 6.66\% compared to the Vanilla method. 
These results indicate that the effectiveness of our approach is not only limited to text classification and can be extended to broader applications.

\section{Related Works}

\textbf{Preference learning.}
Preference learning is about modeling the preference using a set of instances, associated with an order relation \citep{furnkranz2010preference}.
Since it is much easier for humans to make relative judgments, i.e., comparing behaviors as better or worse, preference-based learning becomes an attractive alternative; hence, extensive research has been conducted to address this problem by proposing different techniques to learn from human judgments \citep{biyik2020active, chu2005preference}.
One of the most representative fields that adopt preference-based learning is Reinforcement Learning (RL), to learn RL algorithms from the preferences rather than the explicit design of reward function \citep{wirth2017survey}. 
After the successful scale-up of preference-based learning with deep neural networks \citep{christiano2017deep, lee2021pebble}, this research direction has been extensively explored in other domains such as NLP \citep{stiennon2020learning, ziegler2019fine} and computer vision \citep{kazemi2020preference}, especially focused on the generation tasks, \textit{e.g.,} text summarization and image generation. 
However, preference learning is yet under-explored for classification tasks, despite its great potential to provide complementary and informative training signals via pair-wise comparison of samples.

{\textbf{Learning-to-rank.} 
Preference learning shares a close relationship with learning-to-rank (LTR), a prevalent framework for constructing models or functions to rank objects, as both seek to establish the specific order among samples \citep{hullermeier2008label}.
While preference learning focuses on developing a model to predict preferences between objects, LTR primarily aims to generate ranked lists of items based on their relevance to a given query or context \citep{furnkranz2010preference}. 
Consequently, LTR has become a key component of various information retrieval problems, such as document retrieval and web search \citep{burges2005learning, cao2007learning}. 
Simultaneously, several works have applied LTR to classification tasks; for instance, \citep{chang2020taming} illustrates the efficacy of LTR for multi-label classification. 
Furthermore, \citep{atapour2021rank} transforms classification into LTR and demonstrates its potential for broader classification problems. 
Compared to these works, our work introduces a novel approach to integrating pairwise comparison for generic classification problems through a multi-task learning framework, accompanied by new methods for obtaining pairwise comparisons between samples.}

\textbf{Auxiliary data annotation.}
As the development and deployment of NLP systems are directly affected by the quality of benchmarks, various approaches have been recently explored to construct more effective and robust benchmarks.
For example, one line of works propose to continuously evolve the benchmark to prevent it becomes obsolete or human-aligned by collecting the adversarial samples of the state-of-the-art models \cite{nie2020adversarial, potts2021dynasent} or incorporating human in the data construction loop \cite{kiela2021dynabench, yuan2021synthbio}. 
However, as the collection of new examples is costly, another line of work focuses on finding a better way to annotate the existing benchmarks.
For example, some recent works investigate the alternative labeling method rather than a simple majority voting from the annotation records, to avoid sacrificing the valuable nuances embedded in the annotators’ assessments and their disagreement \cite{fornaciari2021beyond, leonardelli2021agreeing, davani2022dealing}.
Our work suggests a new alternative way for a better annotation of the existing benchmark via preference between pairs of samples.
\section{Conclusion}

In this paper, we introduce task-specific preference signals between pairs of samples as a new and auxiliary data annotation to improve the existing text classification system, which relies on instance-wise annotations. 
To this end, we propose a novel multi-task learning framework, called prefer-to-classify (\name{}), to effectively train the classifier from both task and preference labels, and demonstrate this framework under three different types of preference labels.

\textbf{Acknowledgements.} 
We thank Jongjin Park and Jihoon Tack for providing helpful feedback.
This work was supported by the Institute of Information \& Communications Technology Planning \& Evaluation (IITP) grant
funded by the Korean government (MSIT) (No.2019-0-00075,
Artificial Intelligence Graduate School Program (KAIST); No. 2022-0-00184, Development and Study of AI Technologies to Inexpensively Conform to Evolving Policy on Ethics) and the Engineering Research Center Program through the National Research Foundation of Korea (NRF) funded by the Korean Government (MSIT) (NRF-2018R1A5A1059921).

\bibliography{icml2023_conference}
\bibliographystyle{icml2023}

\appendix
\newpage
\appendix
\onecolumn
\begin{center}{\bf {\LARGE Appendix}}
\end{center}

\begin{center}{\bf {\Large Prefer to Classify: Improving Text Classifiers via Auxiliary Preference Learning}}
\end{center}

\section{Experimental Details}\label{sup:details}
\subsection{Datasets}\label{sup:datasets}

As described in Section \ref{section:setups}, we extensively demonstrate the effectiveness of \name{} on the multiple classification datasets; 10 text classification datasets and 1 image classification dataset. 
For text data, we commonly set the maximum length $L$ as 256 for the tokenization of given data. First, we use the following four text classification datasets in Section \ref{sec:generative}, which do not release the annotation records; hence, the extractive preference labels are not available:

\textbf{CoLA} \citep{warstadt2019neural} is a binary single sentence classification task, where the goal is to predict whether the given English sentence is linguistically valid or not. It is composed of 8.5k training samples and 1k development samples. We remark that the CoLA dataset is part of the popular benchmark, GLUE \citep{wang2019glue}, and the dataset is officially available at \url{https://huggingface.co/datasets/glue}.

\textbf{SMS Spam} \citep{almeida2011contributions} is a public set of SMS-labeled messages that have been collected for mobile phone spam research. It has one collection composed of 5,574 English, real and non-encoded messages, tagged according to being legitimate (ham) or spam. We split the dataset into an 8:1:1 ratio to construct training, validation, and test datasets. SMS Spam is officially available at \url{https://huggingface.co/datasets/sms_spam}.

\textbf{Hate Speech} \citep{fivser2018proceedings} is constructed by extracting the texts from Stormfront, a white supremacist forum. A random set of forum posts have been sampled from several subforums and split into sentences. Those sentences have been manually labeled as containing hate speech or not, according to certain annotation guidelines. Overall, it is composed of 10,703 sentences. We split the dataset into an 8:1:1 ratio to construct training, validation, and test datasets. Hate Speech is officially available at \url{https://huggingface.co/datasets/hate_speech18}.

\textbf{Emotion} \citep{saravia2018carer} is a dataset of English Twitter messages with six basic emotions: sadness (0), joy (1), love (2), anger (3), fear (4), and surprise (5). In the given Emotion dataset, there are 16,000 training, 2,000 validation, and 2,000 test samples. We use the Emotion dataset at \url{https://huggingface.co/datasets/dair-ai/emotion}.

Next, we use the following six text classification datasets, obtained from the following three different sources, which release the annotation records during the construction of the datasets. 
Here, all datasets have the annotation records of 5 different annotators for each data; however, the annotators can be different among data, \textit{i.e.}, there are more than 5 annotators overall. 

\textbf{DynaSent} \citep{potts2021dynasent} is a sentiment classification benchmark with ternary (positive/negative/neutral) sentiments. It is dynamically constructed through multiple iterations of training a classifier model and finding its adversarial samples by involving a human annotator in the loop. In our experiments,  we use the dataset from the first round, \textit{DynaSent-R1}, and the dataset from the second round, \textit{DynaSent-R2}. DynaSent-R1 comprises 80,488 training samples, 3,600 validation samples, and 3,600 test samples, respectively. 
DynaSent-R2 comprises 13,065 training samples, 720 validation samples, and 720 test samples. 
All the validation and test samples are fully balanced between the three classes. 
DynaSent dataset and more details of the dataset are officially available at \url{https://github.com/cgpotts/dynasent}.

\textbf{Standford politeness corpus} \citep{danescu2013polite} is a binary classification benchmark for predicting whether the given sentence is polite or impolite. Since there are two different input domains within this benchmark, we split them into two different datasets: \textit{Polite-Wiki} from Wikipedia, and \textit{Polite-SE} from Stack Exchange, following the original paper \citep{danescu2013polite}.
Here, two classes: polite and impolite, are defined as the top and, respectively, bottom quartile of sentences when sorted by their politeness score. The classes are therefore balanced, with each class consisting
of 1,089 samples for the Wikipedia domain and
1,651 samples for the Stack Exchange domain.
We split each dataset into an 8:1:1 ratio to construct training, validation, and test datasets.
The source data and more details of the dataset are officially available at \url{https://www.cs.cornell.edu/~cristian/Politeness.html}.

\textbf{Offensive agreement dataset} \citep{leonardelli2021agreeing} is a binary classification benchmark for predicting whether the given sentence is offensive or not.
Each sentence is collected from Twitter using Twitter public APIs, based on the hashtags and keywords on three different domains: Covid-19, US Presidential elections
and the Black Lives Matter (BLM) movement.
Remarkably, some of the original samples are not available anymore due to the elimination of tweets from the user side; for example, 10,735 samples are collected initially \citep{leonardelli2021agreeing}, but only 6,513 samples are now available. To address the issue of the reduced number of samples, we slightly modify the dataset to keep the setups of the original paper, \textit{e.g.}, balanced among the classes and domains. 
Specifically, we gather the given splits of the dataset into the unified one and then re-split it as much be balanced as possible. 
This re-constructed dataset has 2,400 training samples, 400 validation samples, and 400 test samples. Also, the ratio between Covid-19, Election, and BLM is 3:3:2.   
The dataset is officially available with the request to authors at \url{https://github.com/dhfbk/annotators-agreement-dataset}.

\textbf{Multi-Genre Natural Language Inference (MultiNLI)} \citep{sener2018multi} is a crowd-sourced collection of 433k sentence pairs annotated with textual entailment information: for a given premise sentence, one should classify whether the given hypothesis sentence is \textit{entailment}, \textit{neutral}, or \textit{contradiction} to the premise (ternary classification).
Since the annotation records are only available with the validation set, we construct the datasets by splitting it into 8:1:1 for training, validation, and test sets. 
This re-constructed dataset has 15,717 training samples, 1,964 validation samples, and 1,966 test samples. 
The source data and more details of the dataset are officially available at \url{https://cims.nyu.edu/~sbowman/multinli}.

Finally, to demonstrate the applicability of \name{} beyond NLP tasks, we use the SUN Attribute dataset \citep{patterson2014sun}:

\textbf{SUN Attribute dataset} \citep{patterson2014sun} is constructed by conducting Amazon's Mechanical Turk crowd-sourcing platform \citep{crowston2012amazon} to annotate the presence of the target attribute in the given image. 
The dataset consists of 14,340 scene images from the SUN dataset \citep{xiao2010sun}, which has 102 scene attributes such as sunny, natural, man-made, etc. In this dataset, the presence of the attribute is measured as an average score of three binary user responses, \textit{i.e.}, it contains the annotation records and hence we use it to construct the extractive preference for our framework.
The source data and more details of the dataset are officially available at \url{https://cs.brown.edu/~gmpatter/sunattributes.html}.

\subsection{Baselines}\label{sup:baselines}

We first introduce some notations for a clear explanation. For each sample $\mathbf{x}$, there are annotation records $n_y(\mathbf{x}) \in \mathbb{N}^{K}$ where $K$ is the number of class and $n_{\tt vote}(\mathbf{x}) = \sum_{y} n_y(\mathbf{x})$ is the number of votes\footnote{All the used datasets commonly have $n_{\tt vote}=5$}. Then, the majority voted target label is obtained by finding the most agreed labels, \textit{i.e.}, $y_{\tt task}(\mathbf{x}) = \arg \max_{y} n_{\tt vote}(\mathbf{x})$, and simply denoted by $y_{\tt task}$. Here, our goal is to train a classifier $f_{\theta}:=W_{\tt task} \circ g_{\phi}$, composed with Transformer-based language model backbone $g_{\phi}$ (\textit{e.g.,} BERT \citep{devlin2019bert}) and a random initialized classification head $W_{\tt task}$, where the prediction for $\mathbf{x}$ is obtained with softmax, \textit{i.e.}, $p(\mathbf{x})={\tt Softmax}\big(f_{\theta}(\mathbf{x})\big)$. 
For the analysis in Figure \ref{fig:analysis}, we only include four baselines with high performance based on the results in Table \ref{table:main_extract}.

\textbf{Vanilla}: as described in Section \ref{sec:prelimiary}, the model $f_{\theta}$ is trained with the following cross-entropy loss:

\begin{equation*}
    \mathcal{L}_{\tt train} = \ell_{\tt xe}(p(\mathbf{x}), y_{\tt task})
\end{equation*}

\textbf{Soft-labeling} \citep{fornaciari2021beyond}: instead of using majority voted label $y_{\tt task}$, it use the soft-labels $q(\mathbf{x})=n_{y}(\mathbf{x}) / n_{\tt vote}(\mathbf{x})$ with a cross entropy loss:

\begin{equation*}
    \mathcal{L}_{\tt train} = \ell_{\tt xe}(p(\mathbf{x}), q(\mathbf{x})) = \sum_{y} - q_{y}(\mathbf{x}) \log p_{y}(\mathbf{x})
\end{equation*}

\textbf{Margin Loss} \citep{sharmanska2016ambiguity}: instead of using majority voted label and cross-entropy loss, it uses the soft-labels $q(\mathbf{x})$ as a margin for the multi-class hinge loss:

\begin{equation*}
    \mathcal{L}_{\tt train} = \sum_{y} \max \{0,  q_{y}(\mathbf{x}) - p_{y}(\mathbf{x})\} 
\end{equation*}

\textbf{Filtering} \citep{leonardelli2021agreeing}: following the setups in the original paper, we exclude the ambiguous samples that have a low agreement between the annotators. 
Specifically, we exclude the samples with $n_{y_{\tt task}} = 3$ since there are 5 annotators for all considered datasets by following \citep{leonardelli2021agreeing}, and use majority voting for the others.

\begin{equation*}
    \mathcal{L}_{\tt train} = \mathbbm{1}[n_{y_{\tt task}}(\boldsymbol{x}) > 3] ~ \ell_{\tt xe}(p(\mathbf{x}), y_{\tt task})
\end{equation*}

\textbf{Weighting} \citep{uma2021learning}: using weighted cross entropy that down-weigh the samples with a low agreement:

\begin{equation*}
    \mathcal{L}_{\tt train} = \mathbf{w}(\mathbf{x}) ~\ell_{\tt xe}(p(\mathbf{x}), y_{\tt task})
\end{equation*}
where $\mathbf{w}(\mathbf{x}) = n_{y_{\tt task}}(\mathbf{x}) / n_{\tt vote}(\mathbf{x})$.

\textbf{Multi-annotator} \citep{davani2022dealing}: instead of aggregating the different annotators' annotation records, it introduces multiple classification heads $W^{(t)}_{\tt task}$ for learning from each annotator's annotation $y^{(t)}_{\tt task}$.
Since each annotator does not annotate all the samples, we simply separate the $n_{\tt vote}(\mathbf{x})$ annotations and train each classification head where $t=1, \dots, n_{\tt vote}$. For the inference of test samples, the ensemble of multiple classification heads is used.

\begin{equation*}
    \mathcal{L}_{\tt train} = \frac{1}{n_{\tt vote}(\mathbf{x})} \sum_{t} ~\ell_{\tt xe}(p^{(t)}(\mathbf{x}), y^{(t)}_{\tt task})
\end{equation*}
where $p^{(t)}(\mathbf{x}) = W^{(t)}_{\tt task} \circ g_{\phi}(\mathbf{x})$.

\textbf{CS-KD} \citep{yun2020regularizing}: for each sample $\mathbf{x}$, the sample $\widehat{\mathbf{x}}$ within the same class, defined by a majority voted label $y_{\tt task}$, is also sample and the consistency regularization is additionally imposed between their prediction with a temperature $\tau$. Following the original paper, we use $\tau = 4$.

\begin{equation*}
    \mathcal{L}_{\tt train} = \ell_{\tt xe}(p(\mathbf{x}), y_{\tt task}) + \ell_{\tt xe}(\widetilde{p}(\mathbf{x}), \widetilde{p}(\widehat{\mathbf{x}})) 
\end{equation*}

where $\widetilde{p}(\mathbf{x}) = {\tt Softmax}(f_{\theta}(\mathbf{x}) / \tau)$.

\textbf{Label Smoothing} \citep{muller2019does}: instead of directly using majority voted label $y_{\tt task}$, it first constructs the soft-label $q(\mathbf{x})$ by subtracting $\tau$ for the class $y_{\tt task}$ and equally distributing it to remaining classes, \textit{i.e.}, $\tau / (K-1)$. We find the best hyper-parameter $\tau$ among $[0.05, 0.1, 0.15]$ using the validation set. Then, this soft label $q(\mathbf{x})$ is used to train the model with a cross-entropy loss:

\begin{equation*}
    \mathcal{L}_{\tt train} = \ell_{\tt xe}(p(\mathbf{x}), q(\mathbf{x})) = \sum_{y} - q_{y}(\mathbf{x}) \log p_{y}(\mathbf{x})
\end{equation*}

\textbf{Max Entropy} \citep{pereyra2017regularizing}: in addition to the cross-entropy loss with the majority voted label $y_{\tt task}$, the regularization loss to increase the entropy of the prediction $p(\mathbf{x})$ is used as a training loss with a hyper-parameter $\lambda$. $\lambda$ is tuned among $[0.1, 0.5, 1.0]$ using the validation set:

\begin{equation*}
    \mathcal{L}_{\tt train} = \ell_{\tt xe}(p(\mathbf{x}), y_{\tt task}) + \lambda \sum_{y} p_{y}(\mathbf{x}) \log p_{y}(\mathbf{x})
\end{equation*}

\renewcommand{\algorithmicrequire}{\textbf{Input:}}
\renewcommand{\algorithmicensure}{\textbf{Output:}}

{\begin{algorithm}[t]
\caption{$\mathtt{Prefer{\text -} to{\text -}Classify~(P2C)~with~extractive~ preference~labels}$}\label{alg:p2c_ext}
\begin{algorithmic}[1]
\REQUIRE Classifier from a pre-trained language model $f_{\theta}$, training dataset $\mathcal{D}$ with preference labels $\{(\mathbf{x}^{0}, \mathbf{x}^{1}, y_{\tt task}, y_{\tt pref})|\mathbf{x}^{0}, \mathbf{x}^{1} \in \mathcal{D} \}$, preference predictors $\{{h}_{\psi^{(t)}}\}_{t=1}^{T}$, mini-batch size $B$, and hyper-parameter $\lambda_{\tt cons}$
\vspace{0.05in}
\hrule
\vspace{0.05in}
\FOR{each iteration}
\STATE Draw a mini-batch $\mathcal{B}=\{(\mathbf{x}_i,y_{{\tt task},i})_{i=1}^{B}\}$ and the corresponding pairs with preference labels $\widetilde{\mathcal{B}}=\{(\widetilde{\mathbf{x}}_i, y_{{\tt pref},i})_{i=1}^{B}\}$ from $\mathcal{D}$ with the inconsistency-based sampling (see Section \ref{sec:p2c})
\STATE Obtain $f_{\theta}(\mathbf{x})$ by forwarding $\mathcal{B}$, then calculate $\mathcal{L}_{\tt multi}$ in Eq. \ref{eq:multi_task}
\STATE Obtain $h_{\psi}(\mathbf{x})$ by forwarding $\mathcal{B}$ and $\widetilde{\mathcal{B}}$, then calculate $\mathcal{L}_{\tt cons}$ in Eq. \ref{eq:consistency_gap}
\STATE Update parameters $\theta$ and $\psi^{(t)}$ to minimize $\mathcal{L}_{\tt train}=\mathcal{L}_{\tt multi} + \lambda_{\tt cons}\mathcal{L}_{\tt cons}$
\ENDFOR
\end{algorithmic}
\end{algorithm}
}
\renewcommand{\algorithmicrequire}{\textbf{Input:}}
\renewcommand{\algorithmicensure}{\textbf{Output:}}

{\begin{algorithm}[t]
\caption{$\mathtt{Prefer{\text -} to{\text -}Classify~(P2C)~ with~subjective/generative~ preference~labels}$}\label{alg:p2c_sub}
\begin{algorithmic}[1]
\REQUIRE Classifier from a pre-trained language model $f_{\theta}$, original training dataset $\mathcal{D}=\{(\mathbf{x}_i,y_i)\}$, collected dataset $\widetilde{\mathcal{D}}$ with preference labels $\{(\mathbf{x}^{0}, \mathbf{x}^{1}, y_{\tt task}, y_{\tt pref})|\mathbf{x}^{0}, \mathbf{x}^{1} \in \mathcal{D}\}$ where $|\widetilde{\mathcal{D}}| = N_{\tt pref}$, preference predictors $\{{h}_{\psi^{(t)}}\}_{t=1}^{T}$, a mini-batch size $B$ and hyper-parameter $\lambda_{\tt cons}$
\vspace{0.05in}
\hrule
\vspace{0.05in}
\FOR{each iteration}
\STATE Draw a mini-batch $\mathcal{B}=\{(\mathbf{x}_i,y_{{\tt task},i})_{i=1}^{B}\}$ from $\mathcal{D}$
\STATE Draw an another mini-batch $\widetilde{\mathcal{B}}=\{(\mathbf{x}_i, \widetilde{\mathbf{x}}_i,y_{{\tt task},i}, y_{{\tt pref},i})_{i=1}^{B}\}$ from $\widetilde{\mathcal{D}}$
\STATE Obtain $f_{\theta}(\mathbf{x})$ by forwarding $\mathcal{B}$, then calculate $\mathcal{L}_{\tt multi}$ in Eq. \ref{eq:multi_task}
\STATE Obtain $h_{\psi}(\mathbf{x})$ by forwarding $\widetilde{\mathcal{B}}$, then calculate $\mathcal{L}_{\tt cons}$ in Eq. \ref{eq:consistency_order}
\STATE Update parameters $\theta$ and $\psi^{(t)}$ to minimize $\mathcal{L}_{\tt train}=\mathcal{L}_{\tt multi} + \lambda_{\tt cons}\mathcal{L}_{\tt cons}$
\ENDFOR
\end{algorithmic}
\end{algorithm}
}

\subsection{Prefer-to-Classify (P2C)}\label{sup:p2c}

In this section, we describe the details of P2C. 
We first note that the details are slightly different between extractive preference learning (Section \ref{section:extract}) and subjective preference learning (Section \ref{section:experiments_subject}) due to the difference in experimental setups between them.
As described in Section \ref{section:experiments}, we commonly use $T=3$ preference heads $\{W_{\tt pref}^{(i)}\}_{i=1}^{T}$ and 2-layer MLPs with ${\tt tanh}$ activation for each $W_{\tt pref}$.
We choose hyper-parameters from a fixed set of candidates based on the validation set; $\lambda_{\tt pref}, \lambda_{\tt div} \in \{1.0, 0.1\}$. Also, we only sample the pair of instances with the same majority voted labels for efficiency.

In the case of learning with extractive preference in Section \ref{section:extract}, we apply the consistency regularization with margin (Eq. \ref{eq:consistency_gap}) by using the difference of annotation as the margin $m$. 
Specifically, we set a margin of class $y$ between two samples $\mathbf{x}^{1}$ and $\mathbf{x}^{0}$ as the difference of their soft-labels $m_{y}=q_{y}(\mathbf{x}^{1}) - q_{y}(\mathbf{x}^{0})$, defined in Section \ref{sup:baselines}. Then, we apply the consistency regularization to all classes $y \in [0,1]^{K}$. In addition, we apply the inconsistency-based sampling for the experiments with extractive preference labels based on the superior experimental results, presented in Section \ref{sup:ablation}.

In the case of learning with subjective and generative preference labels in Section \ref{section:experiments_subject} and \ref{sec:generative}, we apply the consistency regularization without margin (Eq. \ref{eq:consistency_order}) since the explicit degree of preference is not given. 
Also, since the number of pairs with subjective preference labels is limited, we use all of them in training without applying the sampling methods described in Section \ref{sec:p2c}. We introduce the additional mini-batch from these pairs to optimize the model with consistency regularization. The full procedures of P2C with extractive and subjective preference are described in Algorithm \ref{alg:p2c_ext} and \ref{alg:p2c_sub}, respectively.


\newpage

\section{Collection of Preference Labels}\label{sup:interface}

\subsection{More details of preference labels}\label{sup:preference_details}

\begin{figure*}[t]
\begin{center}
    {
    \subfigure[Offensive]
        {
        \includegraphics[width=0.15\textwidth]{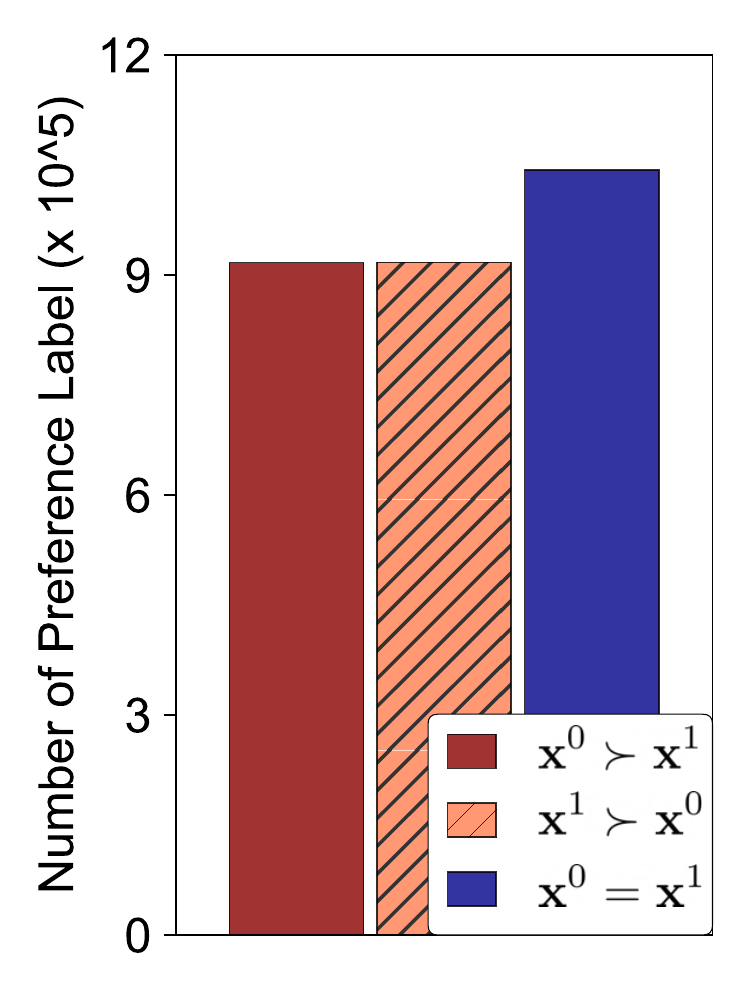}
        }
    \subfigure[Polite-Wiki]
        {
        \includegraphics[width=0.15\textwidth]{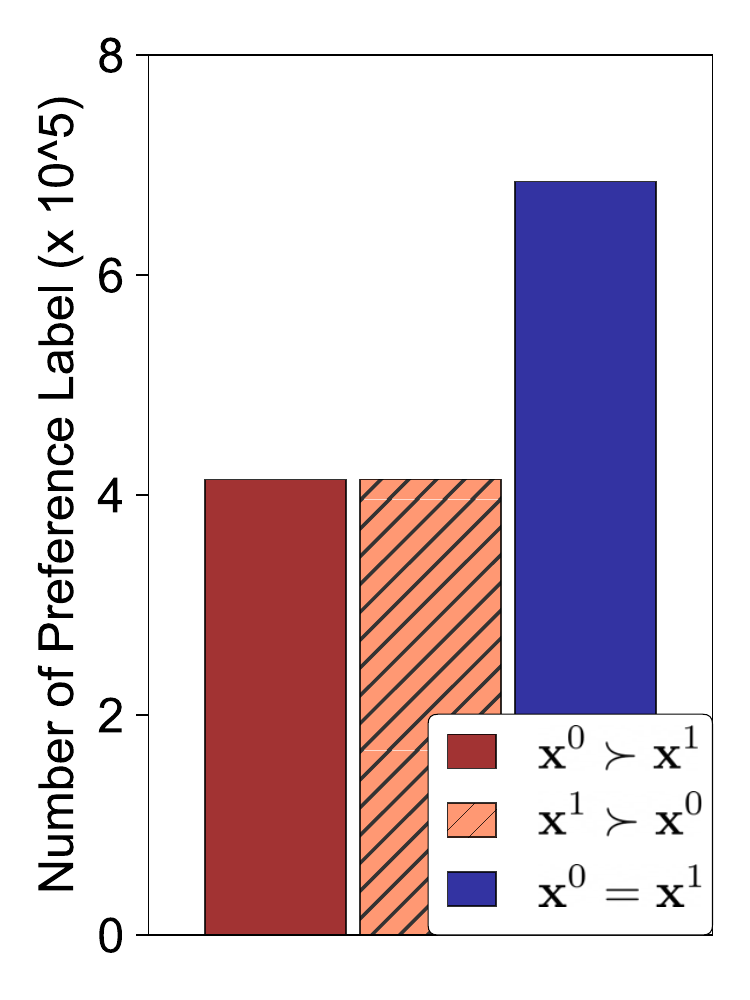}
        }
    \subfigure[Polite-SE]
        {
        \includegraphics[width=0.15\textwidth]{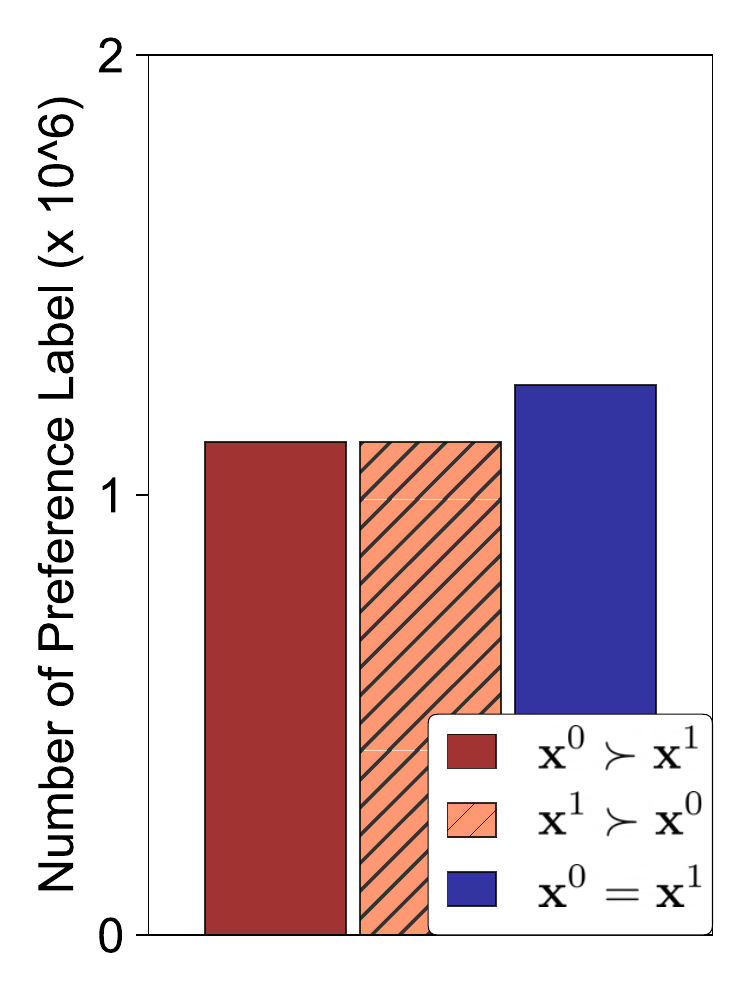}
        }
    \subfigure[MultiNLI]
        {
        \includegraphics[width=0.15\textwidth]{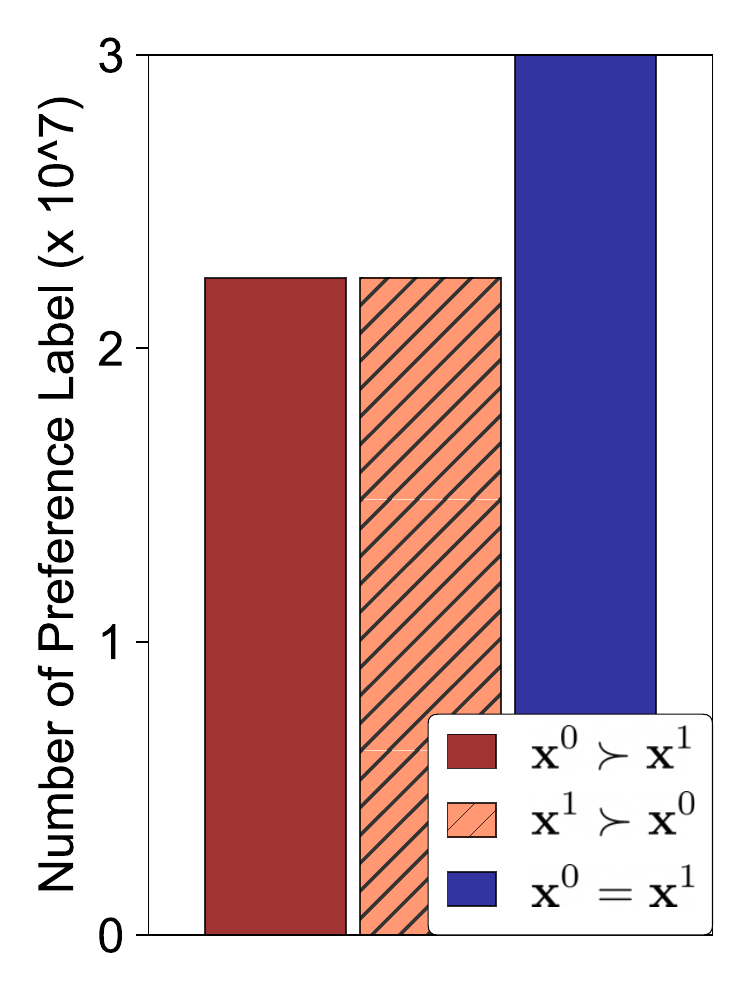}
        }
    \subfigure[DynaSent-R1]
        {
        \includegraphics[width=0.15\textwidth]{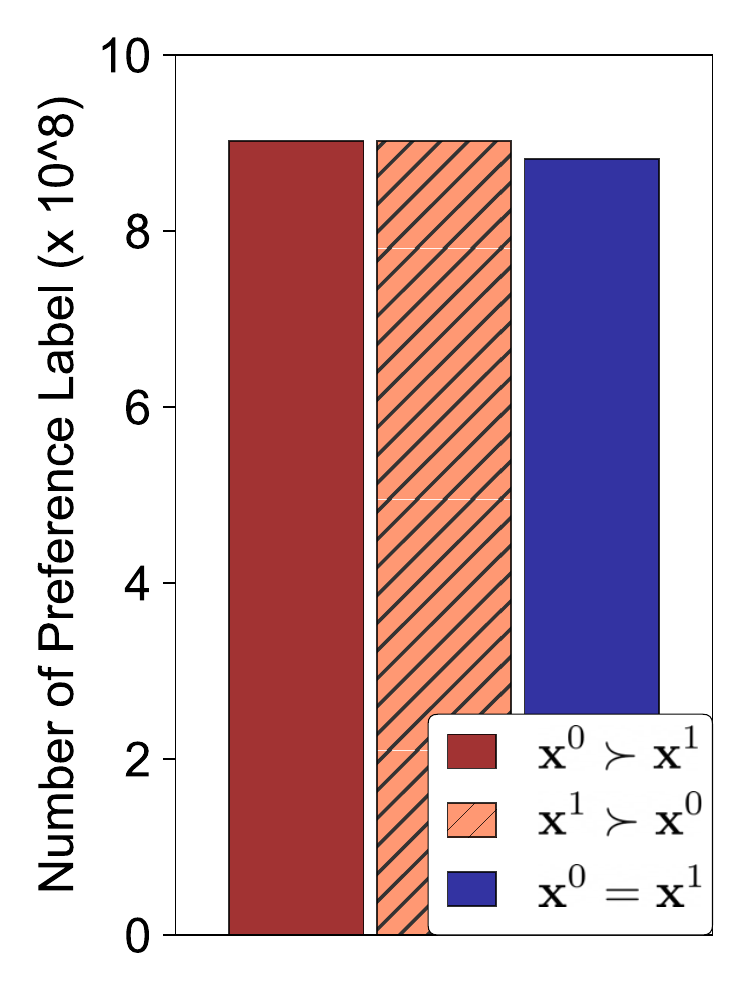}
        }
    \subfigure[DynaSent-R2]
        {
        \includegraphics[width=0.15\textwidth]{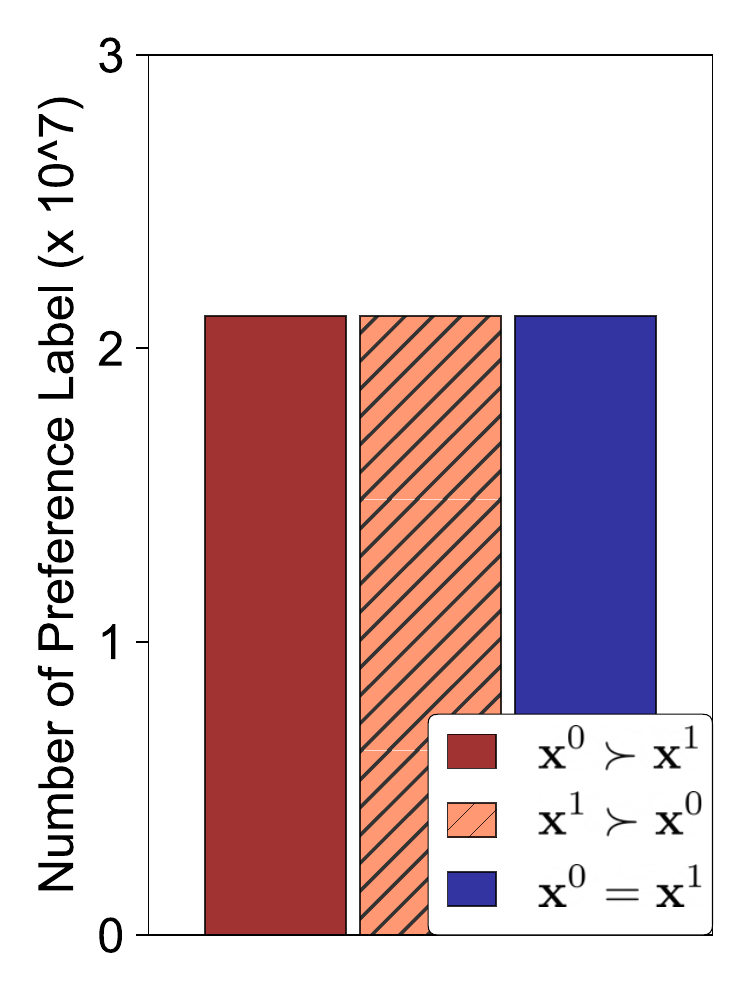}
        } 
    }
\end{center}
\caption{Distribution of the extractive preference labels from the annotation records.}
\label{fig:supp_distb}
\end{figure*}

\textbf{Extractive preference.} For a formal description of the process of collecting extractive preference, we borrow some notations introduced in Section \ref{sup:baselines}. 
As described in Section \ref{sec:preference_collection}, we obtain the extractive preference label $y_{\tt pref}$ by comparing the number of votes $n_{y_{\tt task}}(\mathbf{x})$ with the given task label $y_{\tt task}$: if $n_{y_{\tt task}}(\mathbf{x}^{1}) > n_{y_{\tt task}}(\mathbf{x}^{0})$, then we assign $y_{\tt pref}=1$ where it indicates $\mathbf{x}^{1} \succ \mathbf{x}^{0}$. Similarly, we assign $y_{\tt pref}=0$ when $n_{y_{\tt task}}(\mathbf{x}^{1}) < n_{y_{\tt task}}(\mathbf{x}^{0})$ and $y_{\tt pref}=0.5$ when $n_{y_{\tt task}}(\mathbf{x}^{1}) = n_{y_{\tt task}}(\mathbf{x}^{0})$, respectively.
To reduce the noisy signal and focus on the effective pair, we only compare the samples that have the same majority voted labels, \textit{i.e.}, $y_{\tt task}(\mathbf{x}^{1}) = y_{\tt task}(\mathbf{x}^{0})$.
The resulting distribution of extractive preference labels for each data is presented in Figure \ref{fig:supp_distb}.

\begin{figure*}[t]
\begin{center}
    {
    \subfigure[CoLA]
        {
        \includegraphics[width=0.23\textwidth]{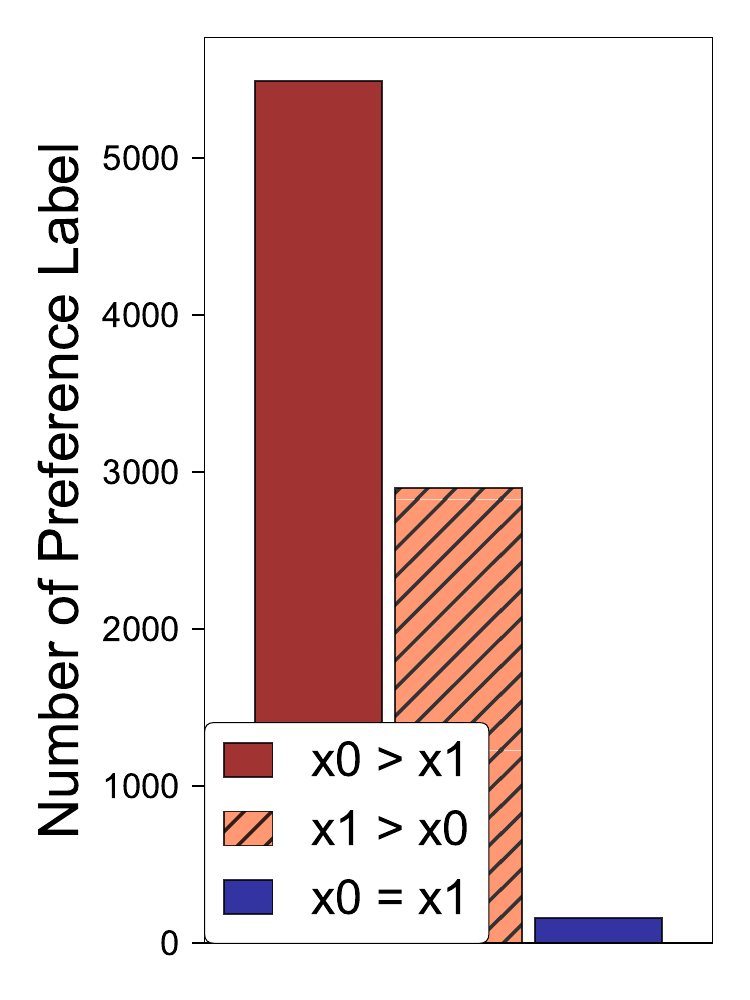}
        }
    \subfigure[SMS Spam]
        {
        \includegraphics[width=0.23\textwidth]{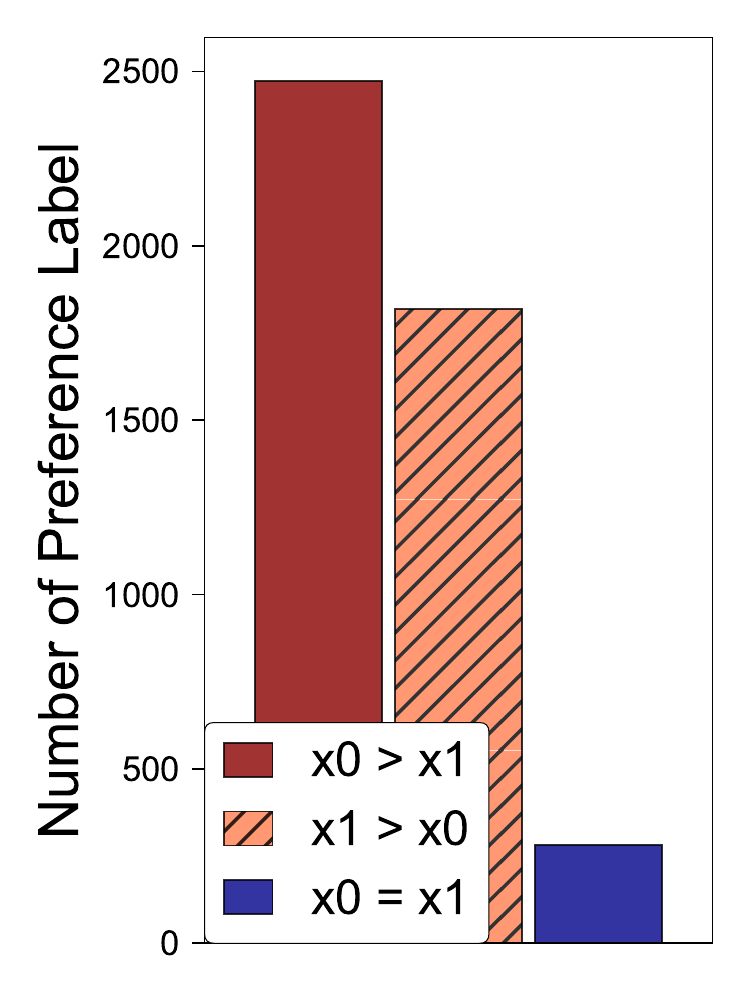}
        }
    \subfigure[Hate Speech]
        {
        \includegraphics[width=0.23\textwidth]{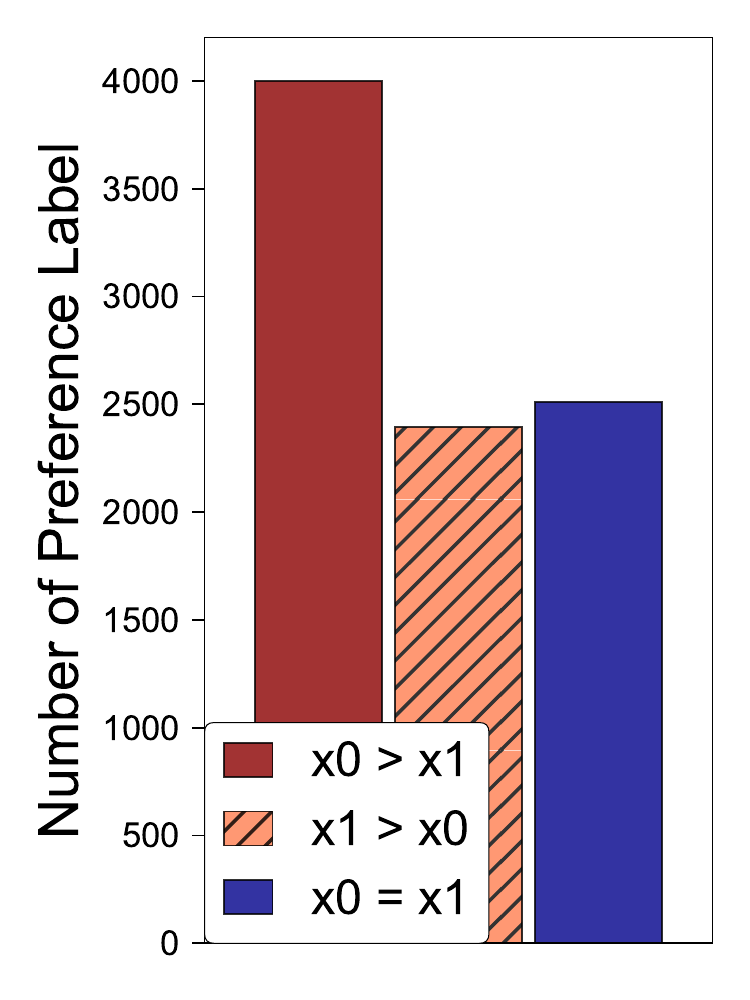}
        } 
    \subfigure[Emotion]
        {
        \includegraphics[width=0.23\textwidth]{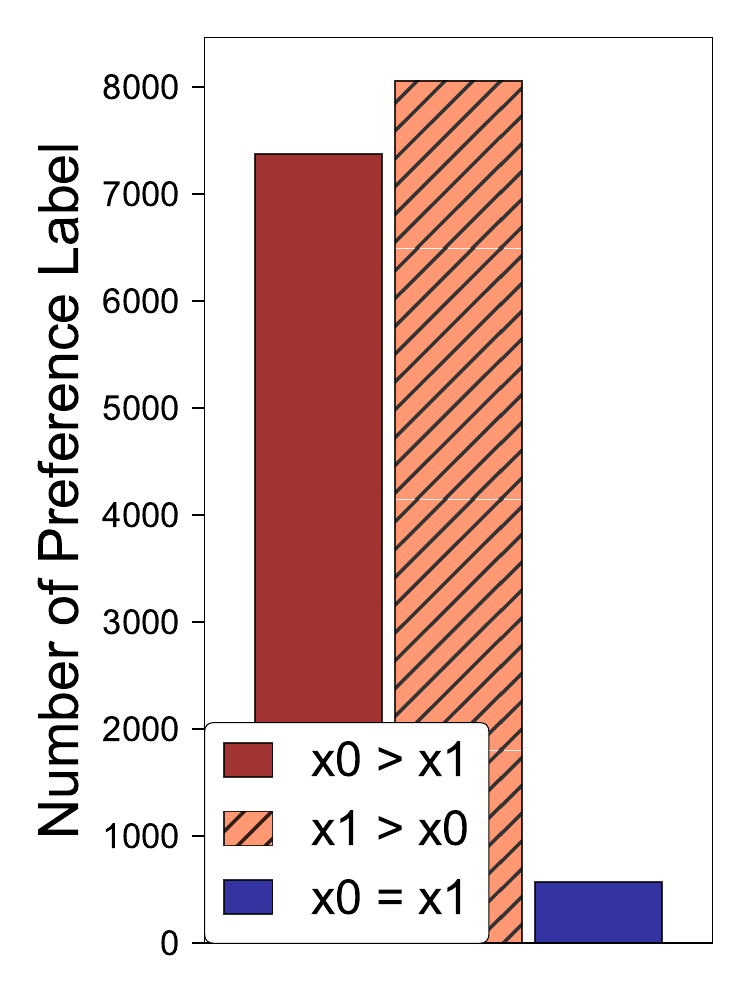}
        } 
    }
\end{center}
\caption{Distribution of the generative preference labels obtained by querying to GPT-3.}
\label{fig:supp_distb_gen22}
\end{figure*}
I\begin{figure*}[t]
\begin{center}
    \includegraphics[width=1.0\textwidth]{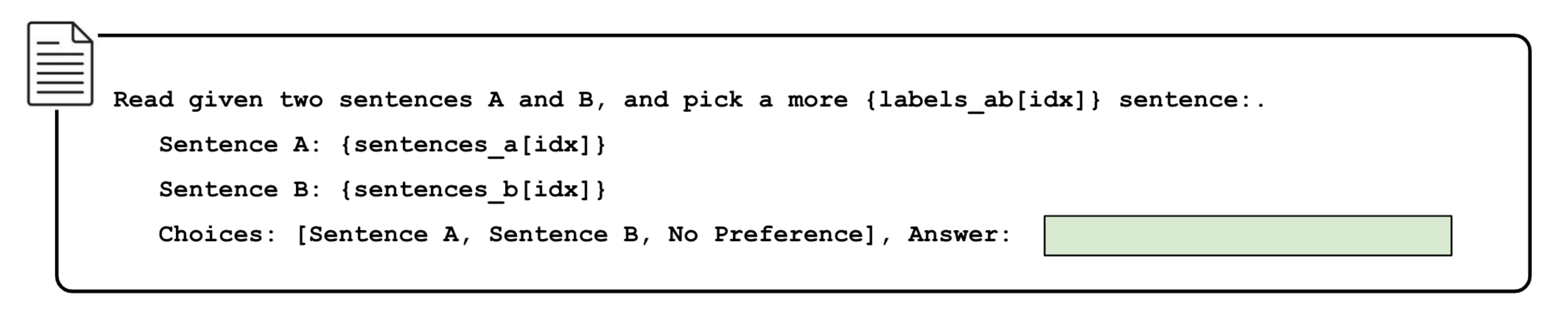}
\end{center}
\vspace{-0.1in}
\caption{Prompt design to collect generative preference labels from GPT-3 \cite{brown2020language}.}
\label{fig:supp_prompt}
\end{figure*}
\begin{figure*}[t]
\begin{center}
    \includegraphics[width=1.0\textwidth]{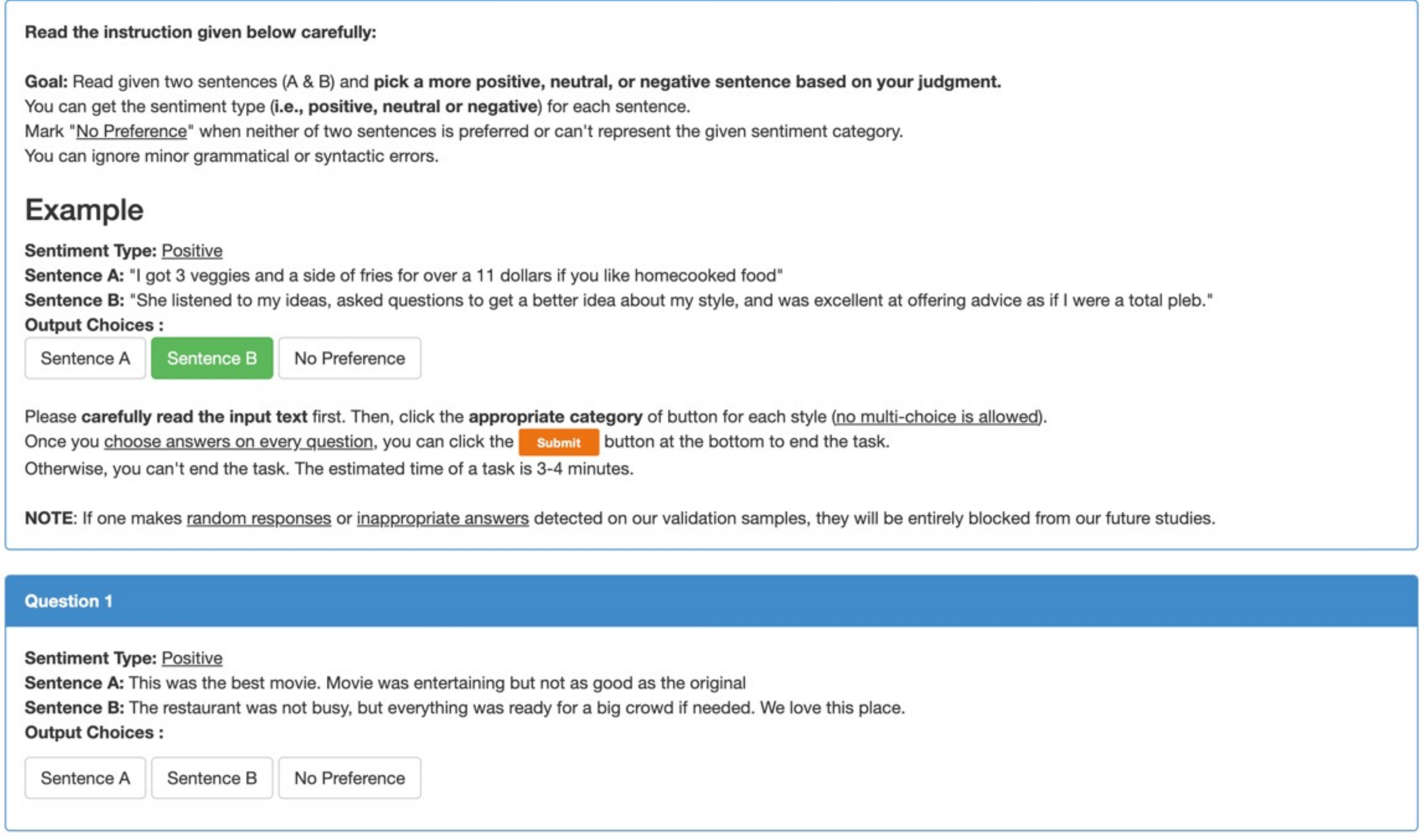}
\end{center}
\vspace{-0.1in}
\caption{Interface to collect subjective preference from crowd workers for sentiment classification (DynaSent-R2 \citep{potts2021dynasent}).}
\label{fig:supp_interface}
\end{figure*}

\textbf{Generative preference.} As we denote in Section \ref{sec:preference_collection}, we collect the generative preference labels by querying the pair of samples to the recent large pre-trained language model, GPT-3 \cite{brown2020language}.
Specifically, we use the officially provided API\footnote{\texttt{text-davinci-003} in \url{https://beta.openai.
com/docs/models/gpt-3}}. To this end, we design our prompt as Figure \ref{fig:supp_prompt}; for $i$th pair of sentences, we provide two sentences along with their task labels. The resulting distribution of generative preference labels for each data is presented in Figure \ref{fig:supp_distb_gen22}. 

\textbf{Subjective preference.} We collect the subjective preference labels based on paired samples from DynaSent-R2 dataset \citep{potts2021dynasent} for the sentiment classification task.
To be specific, we gather the subjective preference of the pairs by asking crowd workers to answer “\textit{which sentence is more positive (neutral, or negative)?}” using Amazon's Mechanical Turk crowd-sourcing platform \citep{crowston2012amazon}. 
Then, each worker should select one of the two sentences or answer “No Preference”. 
Following \citep{nie2020adversarial}, we initially provide each pair of sentences to two crowd workers. 
If two workers give the same preference label, this pair is labeled with that. 
If they disagree, we ask a third crowd worker to break the tie. 
If they still fail to reach a consensus, this pair is labeled with “No Preference”.

Under this procedure, we first gather 1,000 subjective preference labels of randomly selected pairs of sentences. 
Then, we dynamically collect the additional subjective preference labels to maximize the information of collected pairs, motivated by the recent dynamic benchmark constructions \citep{kiela2021dynabench, nie2020adversarial}.
Namely, we first train the model with existing subjective preference labels. 
Then, we find the most informative pairs in the aspect of the trained model, using the disagreement-based sampling introduced in Section \ref{sec:p2c} and query their preference labels in the next stage. 
We select an equal number of pairs for each class to balance the label distribution. 
Overall, starting with 1,000 random pairs, we collect the preference of 2,000 pairs at each round and iterate this procedure for 2 rounds, \textit{i.e.}, a total of 5,000 pairs' subjective preference labels are collected. 

Figure \ref{fig:supp_interface} shows the interface used to collect subjective preference labels from crowd workers for sentiment analysis based on DynaSent-R2 \citep{potts2021dynasent}. 
The top provides the instructions, and then one example is shown. The whole task has 10 items per Human Interface Task (HIT). Workers were paid US\$0.8 per HIT on average, and all workers were paid for their work. 
To improve the quality of collected preference labels, we only hire the Master workers identified as high-performing workers from Amazon’s Mechanical Turk system. 

\subsection{Comparison between different types of preference labels}\label{supp:compare}

\begin{figure*}[t]
\begin{center}
    {
    \subfigure[Preference Label Distribution]
        {
        \includegraphics[width=0.31\textwidth]{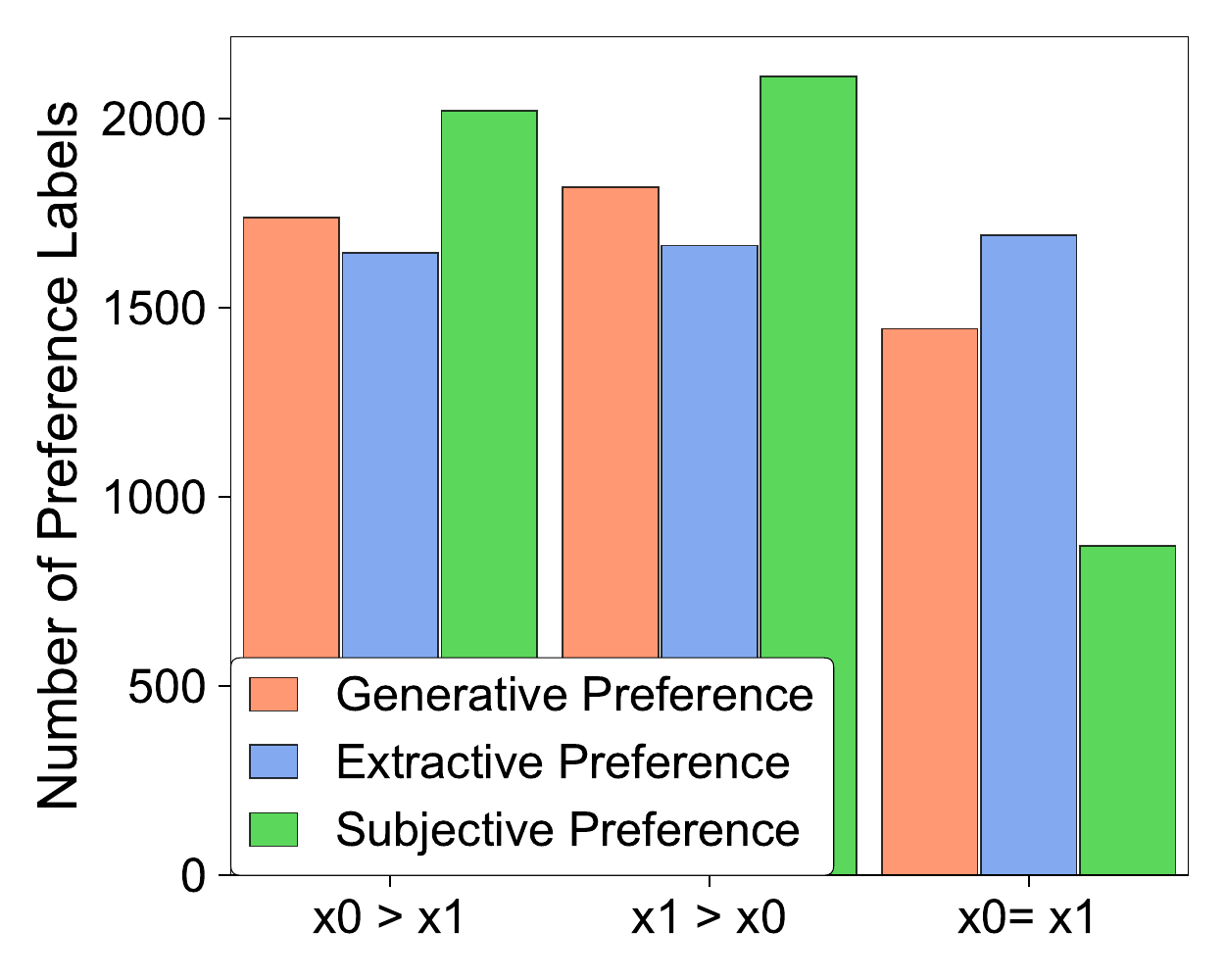}
        \label{sup.fig.8a}
        }
    \subfigure[Overlap between Preference Labels]
        {
        \includegraphics[width=0.31\textwidth]{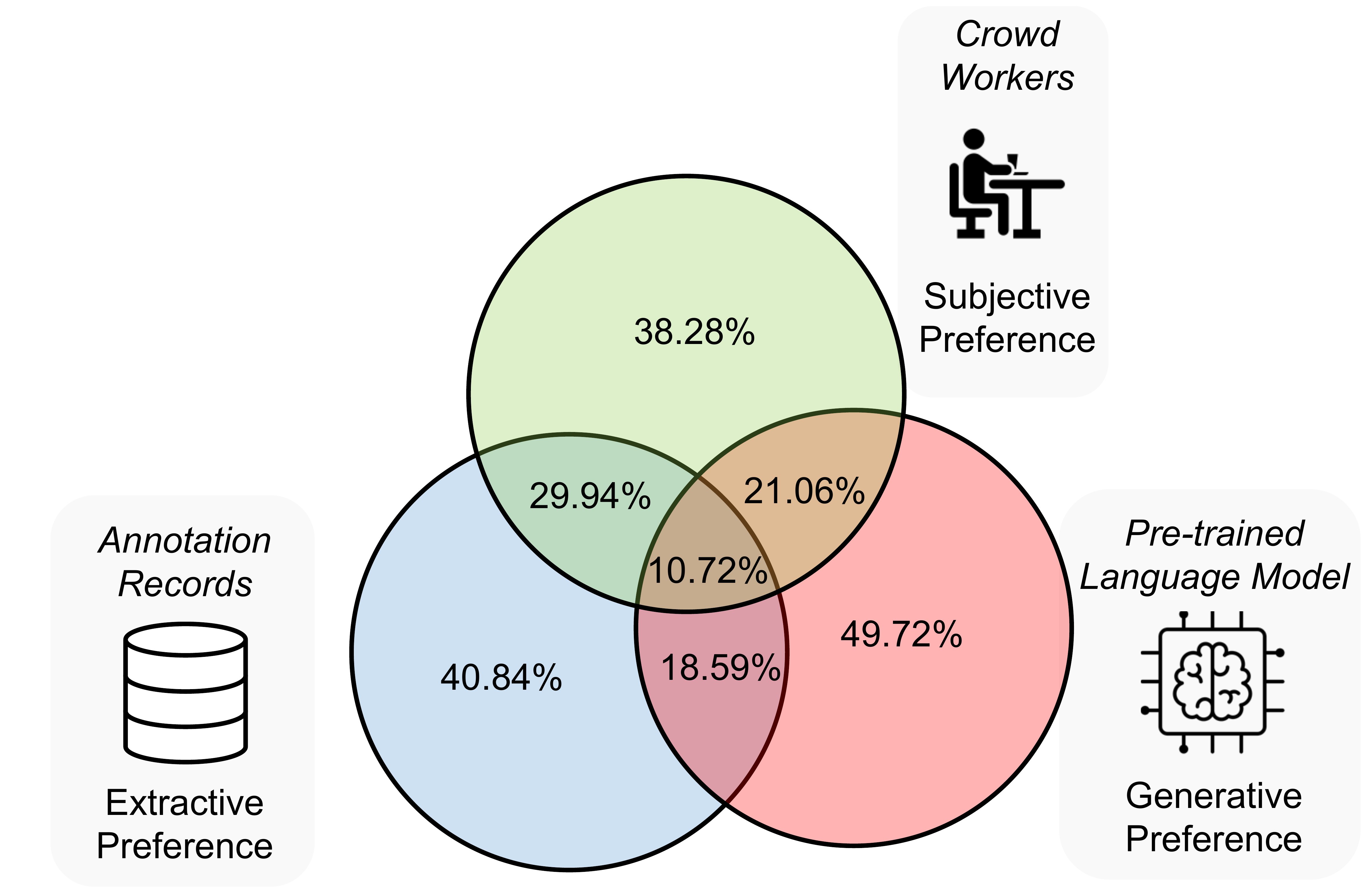}
        \label{sup.fig.8b}
        }
    \subfigure[Results on DynaSent-R2]
        {
        \includegraphics[width=0.31\textwidth]{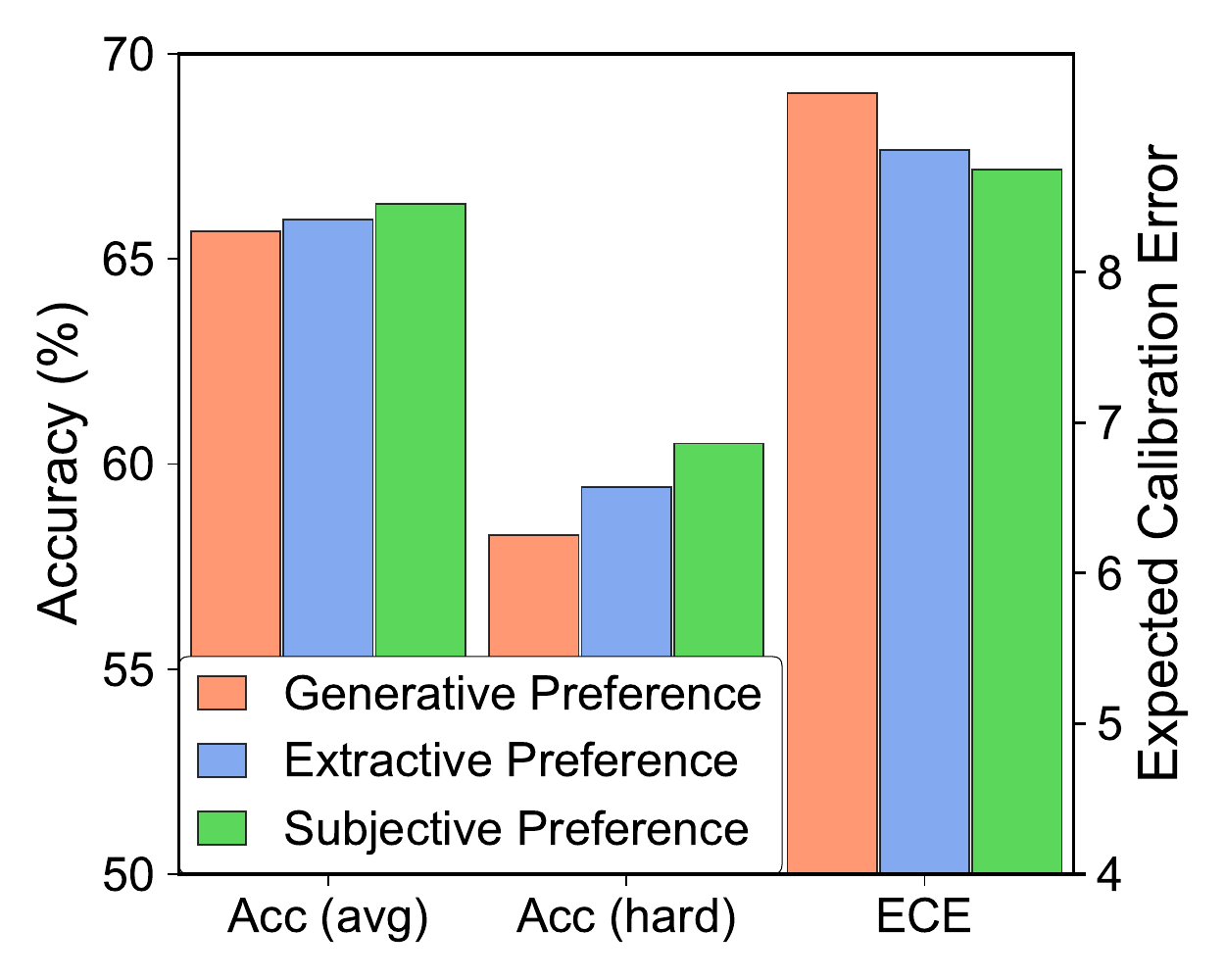}
        \label{sup.fig.8c}
        } 
    }
\end{center}
\caption{Comparison of three different types of preference labels using 5,000 pair of samples on DynaSent-R2. (a) Distribution of each type of preference label. (b) Venn Diagram to denote the similarity (\textit{i.e.}, overlap) between different preference labels. (c) Performance of fine-tuned RoBERTa on the mutually exclusive datasets between all of three different types of preference labels.}
\label{fig:supp_comparison_all_pref}
\end{figure*}

As generative, extractive, and subjective preferences come from different sources of knowledge, they are naturally expected to have different characteristics.
To this end, we first compare the distribution of preference labels with 5,000 pair of samples on DynaSent-R2, which are exactly used to collect the subjective preference labels; as shown in Figure \ref{sup.fig.8a}, they have clearly different label distributions.
This discrepancy is more clearly verified when we measure the coincidence between the preference labels (Figure \ref{sup.fig.8b}); three preference labels output the same label for only 10.72 \% of pairs, while outputting the mutually exclusive one for 19.78 \% of pairs.
To further investigate the effectiveness of each preference label, we fine-tuned RoBERTa model only using those pair of samples with the different preference labels for each method; as shown in Figure \ref{sup.fig.8c}, the subjective preference shows the best performance, while the generative preference shows the worst performance. 
It implies the importance of the quality of preference labels and the effectiveness of generative preference could be from approximating extractive or subjective ones in a cheap way.

\subsection{More examples of preference labels}

In Table \ref{tab:supp_examples}, we present more examples in our extractive, subjective, and generative preference labels on DynaSent-R2 dataset, similar to Table \ref{tab:examples1}. 

\begin{table*}[ht]
\small
\begin{center}
 \caption{More examples in our extractive, subjective, and generative preference labels on DynaSent-R2.}
    \label{tab:supp_examples}
    \begin{adjustbox}{width=1.0\linewidth}
     \begin{tabularx}{\textwidth}{@{}XX@{}}
		\toprule
    \textbf{A}: I noticed when I walked in they looked at me, the eyes of them reflecting.
    & 
    \textbf{B}: I've been to the restaurant a more times and I can understand why this dichotomy may exist.
	\\ 
    \multicolumn{2}{@{}c@{}}{ 
    \colorbox{yellow!30}{Sentiment: \underline{Neutral}, Generative Preference: \textbf{No preference}, Extractive Preference: \textbf{B $\succ$ A}, Subjective Preference: \textbf{B $\succ$ A}}
    }
    \\
    \midrule
    \textbf{A}: The pet clinic was very unprofessional.
    & 
    \textbf{B}: Fast forward to today 2 months later and still I have not received my plates that I paid for and I am driving around on their temp paper plate. I was angry.
	\\ 
    \multicolumn{2}{@{}c@{}}{ 
    \colorbox{yellow!30}{Sentiment: \underline{Negative}, Generative Preference: \textbf{B $\succ$ A}, Extractive Preference: \textbf{No preference}, Subjective Preference: \textbf{No preference}}
    }
    \\
    \midrule

    \textbf{A}: The fresh bread of the bagel available here.
    & 
    \textbf{B}: Since it isn't a big restaurant, to get the attention from the waitress isn't that hard.
	\\ 
    \multicolumn{2}{@{}c@{}}{ 
    \colorbox{yellow!30}{Sentiment: \underline{Positive}, Generative Preference: \textbf{A $\succ$ B}, Extractive Preference: \textbf{B $\succ$ A}, Subjective Preference: \textbf{A $\succ$ B}}
    }
    \\
    \midrule    

    \textbf{A}: I expect everything to turn out well.
    & 
    \textbf{B}: We tried a new place. We couldn't recommend them more highly.
	\\ 
    \multicolumn{2}{@{}c@{}}{ 
    \colorbox{yellow!30}{Sentiment: \underline{Positive}, Generative Preference: \textbf{B $\succ$ A}, Extractive Preference: \textbf{A $\succ$ B}, Subjective Preference: \textbf{A $\succ$ B}}
    }
    \\
    \midrule        
    
    \textbf{A}: But his humor isn't for everyone. I love humor.
    & 
    \textbf{B}: That may have been the norm, but they were above average.
	\\ 
    \multicolumn{2}{@{}c@{}}{ 
    \colorbox{yellow!30}{Sentiment: \underline{Positive}, Generative Preference: \textbf{B $\succ$ A}, Extractive Preference: \textbf{A $\succ$ B}, Subjective Preference: \textbf{No Preference}}
    }
    \\
    \midrule        
    
    \textbf{A}: Management is an embarrassement.
    & 
    \textbf{B}: I saw table of guys feasting on a whole pigs head and having a great time, but it made me pretty sick.
	\\ 
    \multicolumn{2}{@{}c@{}}{ 
    \colorbox{yellow!30}{Sentiment: \underline{Negative}, Generative Preference: \textbf{B $\succ$ A}, Extractive Preference: \textbf{No Preference}, Subjective Preference: \textbf{B $\succ$ A}}
    }
    \\
    \midrule        
    
    \textbf{A}: they put one inside the grocery store.
    & 
    \textbf{B}: We moved here based on reviews and selected South shores with distance of  hours
	\\ 
    \multicolumn{2}{@{}c@{}}{ 
    \colorbox{yellow!30}{Sentiment: \underline{Neutral}, Generative Preference: \textbf{B $\succ$ A}, Extractive Preference: \textbf{No Preference}, Subjective Preference: \textbf{No Preference}}
    }
    \\
    \midrule        
    \end{tabularx}
    \end{adjustbox}
\end{center}
\end{table*}

\newpage

\section{More Ablation Study}\label{sup:ablation}

In this section, we provide more ablation studies on the design choices of P2C. 
Here, all experiments are conducted on DynaSent-R2 \citep{potts2021dynasent} and Offensive \citep{leonardelli2021agreeing} datasets with extractive preference labels, as same as we have done in Section \ref{section:extract}. 
The values and error bars are the mean and standard deviation across five random seeds. 
The results with the chosen design in Section \ref{section:extract} are indicated in \textbf{bold}.

\textbf{Multiple preference heads for preference learning.} 
In Section \ref{sec:p2c}, we introduce multi-preference heads with diversity regularization (Eq. \ref{eq:diversity}) to effectively learn the given preference labels.
To see the effect, we compare it with two different designs for preference heads: 1) single-preference head and 2) multi-preference heads without diversity regularization. 
Remark that the other components, consistency regularization, and inconsistency-based sampling, are still applied to separately verify the effect from different designs of the preference head.
As shown in Table \ref{table:supp_architecture}, one can verify that a single preference head is not enough to exploit the given preference labels fully; hence, the empirical gain is relatively small compared to multi-preference heads. 
Also, it is observable that the proposed regularization is more effective to impose diversity than only relying on random initialization.  

\begin{table*}[ht]
	\begin{center}
	\caption{Effect of different designs for preference head.}
	\vspace{0.1in}
	\label{table:supp_architecture}
    \scalebox{1.0}{
	\begin{tabular}{r|ccc}
		\toprule
		 & { Single-Pref } & { Multi-Pref Heads}  & { Multi-Pref Heads} \\
		{ Dataset} & Head & without diversity & with diversity
		              \\ \midrule
		DynaSent-R2      
		              & {72.22\ms{0.55}}  
		              & {72.75\ms{0.42}}  
		              & \textbf{73.06}\ms{0.31} 
		               \\ 
		Offensive          
		              & {77.08\ms{0.57}}  
		              & {77.25\ms{0.92}}    
		              & \textbf{77.81}\ms{0.21}   
		               \\\bottomrule
	\end{tabular}
    }
    \end{center}
\end{table*}

\textbf{Auxiliary loss for preference learning.} 
As described in Section \ref{sec:p2c}, we use a consistency regularization (Eq. \ref{eq:consistency_order} and \ref{eq:consistency_gap}) between classification and preference learning as an auxiliary loss for learning preference; specifically, consistency regularization with margin (Eq. \ref{eq:consistency_gap}) is used in Section \ref{section:extract}.
To clarify the effectiveness of this regularization, we compare it with 1) consistency regularization without margin (Eq. \ref{eq:consistency_order}). 
We also compare it to 2) soft-labeling, which also uses the annotation records to construct soft-labels instead of the preference and margin. 
Here, we use random sampling instead of inconsistency-based sampling since it is designed explicitly for consistency regularization while using the multi-preference heads.
Table \ref{table:supp_loss} shows the results of these auxiliary losses; although consistency regularization is effective in improving the performance without margin, the gain is smaller than the consistency regularization with margin since the latter utilizes the additional knowledge about the given preference label.
In addition, the result with soft-labeling validates that the gain from our consistency loss is not from the use of the annotation records but from the regularization that imposes the following intuition: \textit{more preferred instance should have higher confidence from the classifier}.

\begin{table*}[ht]
	\begin{center}
	\caption{Effect of different auxiliary losses to learn.}
	\vspace{0.1in}
	\label{table:supp_loss}
    \scalebox{1.0}{
	\begin{tabular}{r|ccc}
		\toprule
		 & Soft & Consistency  & Consistency \\
		Dataset & -labeling & without margin & with margin
		              \\ \midrule
		DynaSent-R2      
		              & {72.29\ms{0.88}}  
		              & {72.40\ms{0.71}}  
		              & \textbf{72.67}\ms{0.89} 
		               \\ 
		Offensive          
		              & {77.04\ms{1.05}}  
		              & {77.54\ms{0.95}}    
		              & \textbf{77.67}\ms{0.99}   
		               \\\bottomrule
	\end{tabular}
    }
    \end{center}
\end{table*}

\textbf{Sampling of pairs for preference learning.} 
To improve the efficiency of preference learning by sampling the informative pairs during the training, we introduce two advanced sampling methods: (1) \textit{disagreement-based} sampling and (2) \textit{inconsistency-based sampling} in Section \ref{sec:p2c}. 
Remark that the other components, consistency regularization with margin and multi-preference heads, are still applied to verify the effect from different sampling methods separately.
In Table \ref{table:supp_sampling}, we compare both sampling methods to random sampling. 
Here, one can verify that both ways are more effective than random sampling, and inconsistency-based sampling is slightly better than disagreement-based sampling. Hence, we commonly used inconsistency-based sampling in Section \ref{section:extract}.

\begin{table*}[ht]
	\begin{center}
	\caption{Effect of different sampling methods.}
	\vspace{0.1in}
	\label{table:supp_sampling}
    \scalebox{1.0}{
	\begin{tabular}{r|ccc}
		\toprule
		Dataset & Random & Disagreement  & Inconsistency \\ \midrule
		DynaSent-R2      
		              & {72.67\ms{0.89}}  
		              & {72.73\ms{0.66}}  
		              & \textbf{73.06}\ms{0.31} 
		               \\ 
		Offensive          
		              & {77.67\ms{0.99}}  
		              & {77.75\ms{1.49}}    
		              & \textbf{77.81}\ms{0.21}   
		               \\\bottomrule
	\end{tabular}
    }
    \end{center}
\end{table*}

{\textbf{Sensitivity to $\mathcal{L}_{\tt div}$.} 
To verify the sensitivity of our method with $\mathcal{L}_{\tt div}$, we conduct the experiments by introducing $\lambda_{\tt div}$, a coefficient of $\mathcal{L}_{\tt div}$, and varying it to investigate its effect. 
In Table \ref{table:supp_lambda}, one can observe that KL divergence does not dominate the entire loss until the certain level of $\lambda_{\tt div}$ including the original value ($\lambda_{\tt div}$=1), but it can diverge with too large value (\textit{e.g.}, $\lambda_{\tt div}=10$). Hence, we recommend using the original value or investigating $\lambda_{\tt div}$ with smaller than 1.

\begin{table*}[ht]
	\begin{center}
	\caption{Effect of diversity regularization between multi-preference heads with $\lambda_{\tt div}$.}
	\vspace{0.1in}
	\label{table:supp_lambda}
    \scalebox{1.0}{
	\begin{tabular}{r|cccc}
		\toprule
		Dataset & $\lambda_{\tt div}=0$ & $\lambda_{\tt div}=1$  & $\lambda_{\tt div}=2$ & $\lambda_{\tt div}=10$\\ \midrule
		DynaSent-R2      
		              & {72.75\ms{0.42}}  
		              & \textbf{73.06}\ms{0.31} 
		               & {71.44\ms{0.68}}  
		                & {57.05\ms{2.14}}  
		               \\ 
		Offensive          
		              & {77.25\ms{0.92}}  
		              & \textbf{77.81}\ms{0.21}   
		              & {75.35\ms{1.03}}    
		              & {65.05\ms{6.70}}    
		               \\\bottomrule
	\end{tabular}
    }
    \end{center}
\end{table*}


\newpage

\section{Additional Experimental Results}\label{sup:rebuttal}

\textbf{Smaller training samples.} 
Here, we validate the effectiveness of P2C with extractive preferences for the smaller training samples. 
Specifically, we control the number of training samples ($N$) of the DynaSent-R2 dataset from $N=250$ to $N=4000$ and compare our method with three representative baselines with high performance: Vanilla, Soft-labeling, and Multi-annotator. 
As shown in Table \ref{table:supp_small}, P2C shows significant improvement, especially when the dataset size is smaller.
We also remark that P2C shows consistent improvement for all cases while other baselines do not.

\begin{table*}[ht]
	\begin{center}
	\caption{Results with the smaller training samples.}
	\vspace{0.1in}
	\label{table:supp_small}
    \scalebox{1.0}{
	\begin{tabular}{r|ccccc}
		\toprule
		Method & $N=250$ & $N=500$ & $N=1000$ & $N=2000$ & $N=4000$ \\ \midrule
		Vanilla      
		                & {54.89\ms{2.46}}  
		                & {60.36}\ms{2.98} 
		                & {63.61\ms{0.92}}  
		                & {66.50\ms{0.76}}
		                & {68.69\ms{1.41}}
		               \\ 
			Soft-labeling      
		                & {57.75\ms{2.35}}  
		                & {60.03}\ms{1.46} 
		                & {62.81\ms{1.45}}  
		                & {66.78\ms{1.16}}
		                & {68.17\ms{1.09}}
		               \\ 
			Multi-annotator      
		                & {57.33\ms{3.23}}  
		                & {61.39}\ms{1.76} 
		                & {63.00\ms{0.87}}  
		                & {66.19\ms{0.84}}
		                & {68.78\ms{1.46}}
		               \\ 
		P2C (Ours)          
		              & \textbf{58.94}\ms{1.16}  
		              & \textbf{61.83}\ms{1.15}   
		              & \textbf{64.13}\ms{1.04}    
		              & \textbf{67.72}\ms{0.46}    
		              & \textbf{69.83}\ms{0.64}    
		               \\\bottomrule
	\end{tabular}
    }
    \end{center}
\end{table*}

\textbf{Compatibility with other types of models.}
While we have previously used a model built over RoBERTa-base \citep{liu2019roberta}, the proposed \name{} is not limited to the specific model. 
To verify this, we conduct additional experiments based on DynaSent-R2 with extractive preference labels from the annotation records. 
As shown in Table \ref{table:supp_other_lm}, the proposed \name{} consistently improves the test accuracy of classifiers of other language models: BERT-base \citep{devlin2019bert}, ALBERT-base \citep{lan2020albert}, and RoBERTa-large. 

\begin{table*}[ht]
	\begin{center}
	\caption{Results with other types of language models.}
	\vspace{0.1in}
	\label{table:supp_other_lm}
    \scalebox{1.0}{
	\begin{tabular}{r|ccc}
		\toprule
		Method & BERT-base & ALBERT-base & RoBERTa-large \\ \midrule
		Vanilla      
		               & {67.26\ms{1.15}} 
		              & {62.72}\ms{0.73} 
		                & {75.62\ms{0.60}}  
		               \\ 
		P2C (Ours)         
		               & {68.26\ms{0.56}} 
		              & {65.00}\ms{1.13}   
		               & {77.71\ms{0.36}}  
		               \\\bottomrule
	\end{tabular}
    }
    \end{center}
\end{table*}

\end{document}